\documentclass[]{fairmeta}

\usepackage{silence}
\usepackage[misc]{ifsym}
\newcommand{\methodname}{SAM2Matting\xspace}

\title{\vspace{-1pt}{\fontsize{19pt}{22.5pt}\selectfont \methodname: Generalized Image and Video Matting}}

\author[1]{Ruiqi Shen}
\author[1]{Guangquan Jie}
\author[2]{Chang Liu}
\author[1]{Henghui Ding}

\affiliation[1]{Fudan University}
\affiliation[2]{Shanghai University of Finance and Economics}

\usepackage[utf8]{inputenc}
\usepackage[T1]{fontenc}
\usepackage{color}
\usepackage{epsfig}
\usepackage{graphicx}
\usepackage{booktabs}
\usepackage{tabularx}
\usepackage{multirow}
\usepackage{amsmath,amsfonts,amssymb}
\usepackage{bm}
\usepackage{nicefrac}
\usepackage{microtype}
\usepackage{xspace}
\usepackage{url}
\usepackage[table]{xcolor}
\usepackage{caption}
\usepackage{subcaption}
\usepackage{makecell}
\usepackage{cleveref}
\usepackage{stfloats}
\usepackage{xcolor}
\usepackage{wrapfig}
\definecolor{citecolor}{HTML}{0071BC}
\usepackage{hyperref}
\hypersetup{colorlinks=true, citecolor=citecolor, linkcolor=citecolor}
\usepackage{bbm}

\crefname{figure}{Fig.}{Figs.}
\crefname{table}{Tab.}{Tabs.}

\usepackage[round]{natbib}

\definecolor{defaultColor}{RGB}{230, 230, 250}
\newcommand{\pub}[1]{\color{gray}{\raisebox{0.25ex}{\scalebox{0.8}{[{#1}]}}}}
\definecolor{rred}{RGB}{245, 152, 153}
\definecolor{oorange}{RGB}{253, 205, 154}
\definecolor{yyellow}{RGB}{255, 240, 180}
\definecolor{ggreen}{RGB}{200, 230, 200}

\abstract{
    Despite impressive advances in image matting, video matting remains challenging due to the inherent gap between high-level tracking, which requires frame-wise understanding, and low-level matting, which focuses on extremely fine-grained details. Existing methods attempt this with expensive and narrowly-scoped video matting datasets, which may limit out-of-domain generalization and compromise tracking robustness. We rethink the paradigm with \textit{\methodname}, a tracker-to-matting framework that advances VOS trackers to high-fidelity video matting. Specifically, it decouples the task by enhancing a foundational tracker (e.g., SAM2, SAM3) with a region-proposal bridge and dedicated matting heads, enabling the uncompromised tracker to handle temporal consistency while the matting components resolve fine-grained details. Notably, despite being trained only on images, \methodname establishes new state-of-the-art performance on video matting, supports diverse prompt types, maintains strong temporal consistency, and demonstrates robust generalization across both human-centric and in-the-wild scenarios.
}

\metadata[Website]{\url{https://henghuiding.com/SAM2Matting}}
\metadata[Code]{\url{https://github.com/FudanCVL/SAM2Matting}}
\metadata[Email]{henghui.ding@gmail.com}

\begin{document}
\maketitle

\section{Introduction}

\label{sec:intro}

Matting aims to separate the foreground target from its background by predicting a pixel-level alpha matte. Evaluated by transparency values, it is generally considered as a fundamental low-level vision task ~\citep{xu2017deep, ding2022deep, hou2019context, Liu_2021_WACV, lin2021real, yao2024matte}.

\begin{figure*}[b!]
  \centering
  \includegraphics[width=\linewidth]{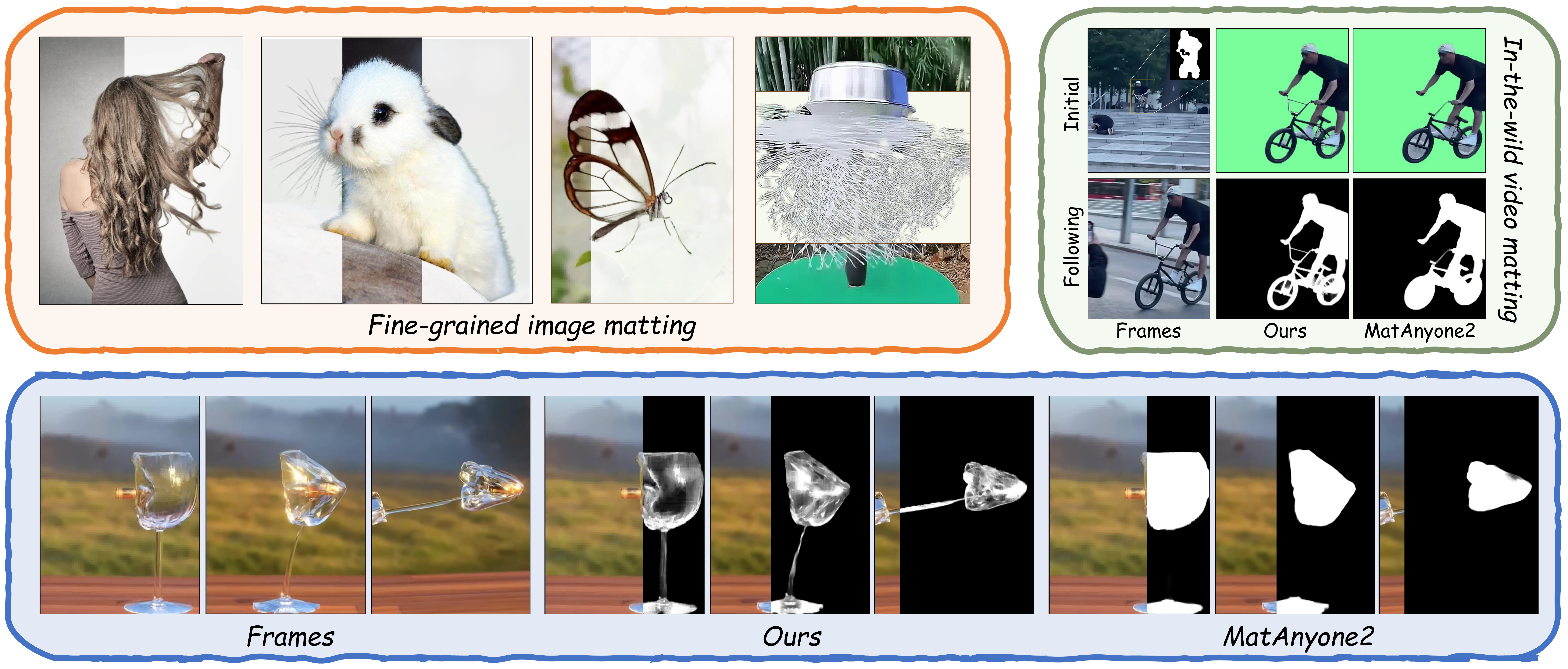}
  \vspace{-0.25in}
  \captionof{figure}{\methodname enables fine-grained image and video matting across both human and in-the-wild scenarios.}
  \label{fig:figure1}
\end{figure*}

When extending to videos, it is common practice to require explicit target specification, typically an initial-frame mask, to disambiguate the target and enable consistent tracking across subsequent frames \citep{park2023mask, huynh2024maggie, zhang2025object, yang2025matanyone, yang2025matanyone2}. Consequently, video matting faces a fundamental trade-off: it demands both high-level semantic understanding to robustly track the target as in Video Object Segmentation (VOS) and low-level fine-grained perception to capture extremely intricate details as in image matting. To bridge this gap, recent approaches rely heavily on video matting datasets \citep{lin2021real, huang2023end, wang2023video, huynh2024maggie, yang2025matanyone, yang2025matanyone2, lim2026videomama, zhang2025object}. However, the prohibitive cost of annotating pixel-level alpha values across frames restricts these video matting datasets to limited scales and narrowly-focused domains, primarily human-centric \citep{lin2021real, huynh2024maggie, yang2025matanyone, yang2025matanyone2}, leaving them insufficient for representing rich real-world dynamics compared with large-scale video segmentation datasets \citep{ravi2024sam, ding2025mosev2, ding2023mose, MeViS, MeViSv2}. As a result, training a model from scratch on such constrained data fails to establish robust tracking capabilities, while fine-tuning a pretrained VOS model on them compromises the model's original tracking robustness (see Figure \ref{fig:video-ft}). These challenges motivate our core question: \textit{must we depend on these heavily-annotated and still narrowly-scoped video matting datasets?}

Rethinking this paradigm, we argue that video matting inherently combines two distinct sub-tasks: high-level tracking for temporal consistency, which is already well-addressed by established VOS models trained at scale, and low-level matting for intricate spatial details, which is comprehensively captured by diverse image matting datasets. We therefore propose \textit{\methodname}, a decoupled tracker-\textit{to}-matting (\textit{hence 2}) framework that brings established VOS trackers (e.g., SAM2, SAM3) to high-fidelity video matting by preserving their high-level robustness while training dedicated components on rich and diverse image matting data for low-level alpha estimation. Our framework seamlessly integrates robust tracking, matte detection, and alpha refinement into a unified pipeline: multi-source priors provide guidance to identify matting-critical regions, where alpha mattes are generated and progressively refined for fine-grained estimation.

Specifically, our matting components take image-level cues, including multi-scale image features, together with mask-level tracking priors as inputs. To bridge high-level tracking to fine-grained matting, we propose an ROI Detector that identifies \textit{matting-critical} regions, i.e. regions with fine-grained details or semi-transparency, while rectifying tracking inaccuracies. This differs from prior methods, which rely on rule-based morphological operations for identifying these regions \citep{sharma2020alphanet, zhou2021semantic, yao2024matte} or directly use the raw mask for matting \citep{yu2021mask, park2023mask, li2024matting, yang2025matanyone, yang2025matanyone2}. Subsequently, a Progressive Alpha Predictor generates and refines the mattes within the identified ROIs through a multi-scale cascade, supervised at all intermediate scales \citep{cheng2022masked}. Tailored losses are introduced to smooth transitions and preserve matte integrity, ensuring high-fidelity results.

We provide three variants of \methodname based on different VOS trackers: SAM2.1-Tiny, SAM2.1-Base+ \citep{ravi2024sam} and SAM3 \citep{carion2025sam}. Comprehensive experiments show that \methodname achieves state-of-the-art (SOTA) performance on both image and video matting, with video matting evaluated in a strictly \textbf{zero-shot} manner. Extensive in-the-wild results further demonstrate its strong generalization to open-world scenarios with rapid motion, complex backgrounds, and target attachments (e.g., man riding a bicycle). Moreover, our matting components are lightweight and efficient, enabling the SAM2.1-Tiny variant to run at 40 FPS on a 200-frame 1080p video using less than 5GB GPU memory.

Our main contributions are summarized as follows:

\begin{itemize}
    \item We present \textit{\methodname}, a new matting paradigm that decouples video matting into high-level tracking and low-level matting for optimal performance when integrated. Specifically, multi-source priors are used to identify matting-critical regions, followed by progressive alpha flows for cascaded refinement.

    \item \methodname achieves new SOTA performance on video matting in a \textit{zero-shot} way, eliminating costly and painstaking annotations while generalizing robustly to both human-centric and in-the-wild scenarios.

    \item \methodname seamlessly adapts to different foundational trackers. We open-source three variants based on SAM2.1-Tiny, SAM2.1-Base+, and SAM3, supporting diverse prompt types including mask, point, box, and text. Crucially, \methodname introduces minimal FPS and VRAM overhead over the trackers, while delivering stable matting performance over extended and challenging real-world videos.
\end{itemize}

\section{Related Work}

\subsection{Image Matting}

Given an image \( I \),  matting aims to separate a foreground target \( F \) from its background \( B \) with pixel-level precision by predicting an alpha matte \( \alpha \) , letting \( I = \alpha F + (1 - \alpha) B \). Previous deep image matting methods can be broadly categorized into two paradigms. Automatic (or auxiliary-free) matting directly mats all objects within the image \citep{zhang2019late, deora2021salient, li2021privacy, yu2021cascade, dai2022enabling}, requiring no additional input beyond the image itself. However, this paradigm struggles in real-world complex scenes where target identification becomes ambiguous \citep{huynh2024maggie, yang2025matanyone}. In contrast, prompt-based matting requires the target to be specified, with earlier methods relying on trimaps~\citep{hou2019context,yu2021high,park2022matteformer,li2024drip,hu2024diffusion} for precise guidance. Recently, some methods have relaxed the trimap requirement to coarser forms such as points, scribbles, boxes, or directly use the masks, achieving promising results~\citep{yang2020smart, park2023mask, yao2024matte, li2024matting, ding2022deep, tan2016novel, wei2021improved, liu2025enhancing}.

\subsection{Video Matting}

A few early video matting approaches \citep{lin2022robust, li2023videomatt, li2024vmformer} are target-agnostic, estimating alpha mattes for all visible objects across the sequence. However, this setting becomes ambiguous in real-world videos where targets may frequently enter and exit the scene \citep{huang2023end, huynh2024maggie, yang2025matanyone}. Recent methods therefore require explicit target specification: earlier approaches use per-frame or initial-frame trimaps \citep{zhang2021attention, sun2021deep, seong2022one}, while newer ones replace trimaps with masks \citep{huang2023end, wang2023video, zhang2025object, yang2025matanyone, yang2025matanyone2, lim2026videomama} and adapt VOS priors through training on video matting data. However, their performance remain limited by the scale and diversity of existing video matting datasets, which mostly require exhaustive and labor-intensive annotations and are predominantly human-centric \citep{lin2021real, huynh2024maggie, yang2025matanyone}. Very recently, generative pipelines have emerged for automatically synthesizing video matting data, but they either remain human-centric \citep{yang2025matanyone2} or scale through pseudo-labeling existing large-scale VOS benchmarks \citep{lim2026videomama}. This motivates us to ask whether robust and generalizable video matting across both human-centric and in-the-wild scenarios can be achieved zero-shot.

\section{Method}

\subsection{Overview}

\begin{figure*}[!t]
  \centering
  \includegraphics[width=1.0\textwidth]{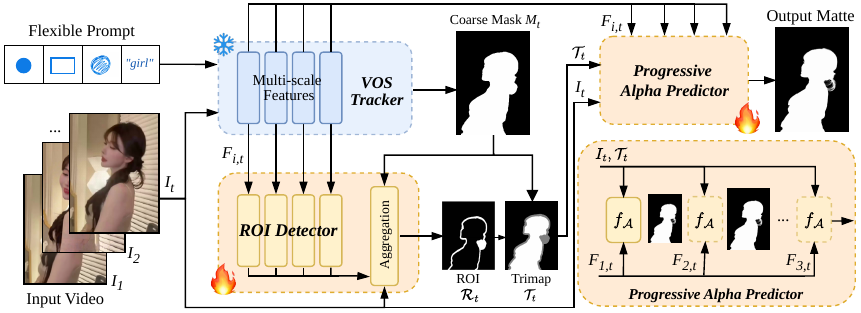}
  \vspace{-0.25in}
  \caption{\textbf{Overview of \methodname.} \methodname adopts a decoupled design that leverages a VOS tracker for high-level tracking and dedicated components for low-level matting. The Region of Interest (ROI) Detector identifies matting-critical regions by integrating image-level and mask-level priors. The Progressive Alpha Predictor then generates and refines the alpha mattes across scales through cascaded refinement.}
  \label{fig:figure2}
\end{figure*}

Figure \ref{fig:figure2} illustrates \methodname, a generalized framework for image and video matting that decouples high-level tracking from dedicated low-level matting components. Specifically, a VOS tracker provides a temporally-consistent target mask for each frame. Given the mask and multi-scale image features, an ROI Detector identifies matting-critical regions with fine-grained details or semi-transparency. A Progressive Alpha Predictor then iteratively produces and refines the matte through a coarse-to-fine cascade, with intermediate mattes supervised at each scale to progressively capture finer details.

\subsection{Regions of Interest (ROI) Detector}

\label{sec:ROI detector}
\begin{figure*}[!t]
  \centering
  \includegraphics[width=1.0\columnwidth]{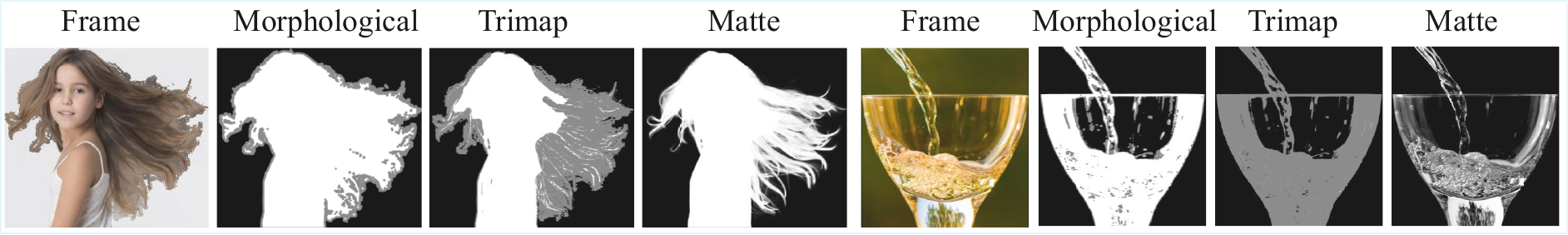}
  \vspace{-0.28in}
  \caption{Simple morphological operations (dilation \& erosion) on the foreground mask produce coarse, uniform boundaries (``Morphological''), which fail to capture complex regions in the ground-truth trimap (``Trimap''), such as the fine structures in flowing hair (\textit{left}) and translucent water cup (\textit{right}).}
  \label{fig:figure3}
\end{figure*}

Typical matting frameworks generate regions of interest (ROI), or ``unknown'' regions, as an intermediate step to concentrate the model on the most relevant areas for alpha estimation. However, conventional approaches derive these regions in simple rule-based ways. One common strategy employs morphological operations on the mask (e.g. dilation and erosion) \citep{sharma2020alphanet, zhou2021semantic, yao2024matte}, which implicitly assumes uniform boundary importance (Figure \ref{fig:figure3}). Alternatively, other methods directly use the raw mask as the ROI \citep{yu2021mask, park2023mask, li2024matting}. Both approaches are prone to overlook fine details within complex structures or include definite foreground regions that do not require matting.

To overcome these limitations, we introduce an ROI Detector to precisely identify matting-relevant regions in each frame. We reformulate ROI detection as a pixel-wise binary classification task, where positive pixels denote matting-critical areas. The detector integrates diverse priors, including the VOS mask \(M\), the current frame \(I\), and the multi-scale image features \(F\). Since features at different resolutions capture different levels of semantics and details, we predict an ROI logit map at each scale \( i \in \{1, \dots, n\} \).

Specifically, at frame \(t\), for each scale \(i \in \{1, \dots, n\}\), a scale-specific convolutional ROI head \(f_{\mathrm{R},i}(\cdot)\) estimates an ROI logit map \(L_{t,i} \in \mathbb{R}^{H_i \times W_i}\) from the image feature \(F_{t,i} \in \mathbb{R}^{C_i \times H_i \times W_i}\), the resized frame \(I_{t,i} \in \mathbb{R}^{3 \times H_i \times W_i}\), and the resized mask \(M_{t,i} \in \{0,1\}^{H_i \times W_i}\):
\begin{equation}
L_{t,i} = f_{\mathrm{R},i}\left(F_{t,i}, M_{t,i}, I_{t,i}\right).
\end{equation}
The logit maps at different scales are then aggregated by a hierarchical convolutional network \( f_{\varphi}(\cdot) \), which integrates global context with structural details, producing \( L_{t} \in \mathbb{R}^{H \times W} \) for frame \( t \):
\begin{equation}
L_t = f_{\varphi} \left( \left[U\!\left( L_{t,1} \right), U\!\left( L_{t,2} \right), \cdots, U\!\left( L_{t,n} \right) \right] \right),
\end{equation}
where \( U(\cdot) \) and $[\cdot]$ denote upsampling and concatenation, respectively. The final ROI prediction \( \mathcal{R}_t \in \{ 0, 1 \}^{H \times W} \) for frame \( t \) is then obtained by applying a sigmoid activation $\sigma$ to \( L_t \) and thresholding it with $ \theta $:
\begin{equation}
\mathcal{R}_t = \mathbbm{1}\left[\sigma(L_t) \geq \theta\right].
\end{equation}
During training, we supervise $L_t$ with focal and smoothness losses (see Section \ref{sec:optimization_strategies}), encouraging $\mathcal{R}_t$ to precisely capture regions with intricate details or semi-transparency that require matting.

\subsection{Pseudo-Trimap Generation}

The predicted ROI $\mathcal{R}_t$ identifies regions requiring fine-grained matting. To preserve structural integrity and provide explicit foreground-background separation, we construct a pseudo-trimap $ \mathcal{T}_t \in \{ 0, 0.5, 1 \}^{H \times W} $ by assigning the VOS mask \( M_t  \in \{ 0, 1 \}^{H \times W} \) to definite foreground and background, while designating $\mathcal{R}_t$ as the unknown region. Specifically, the pseudo-trimap \(\mathcal{T}_t\) for frame $t$ is formulated as:

\begin{equation}
\mathcal{T}_{t,(h,w)} =
\begin{cases}
M_{t,(h,w)}, & \text{if } \mathcal{R}_{t,(h,w)} = 0, \\
0.5, & \text{if } \mathcal{R}_{t,(h,w)} = 1.
\end{cases}
\end{equation}

The resulting \(\mathcal{T}_t\) assigns labels for definite foreground (1), definite background (0), and unknown regions (0.5), providing pixel-level spatial priors for subsequent alpha estimation.

\subsection{Progressive Alpha Predictor}

Unlike the ROI Detector's parallel processing, the Progressive Alpha Predictor treats alpha estimation as a sequential refinement process. It adopts a coarse-to-fine strategy, where each intermediate alpha prediction is passed to the subsequent scale as guidance for refinement. Specifically, at frame $t$, the composite input $X_{t,i}$ at scale $i$ concatenates the image feature $F_{t,i}$, the resized frame $I_{t,i}$, the resized pseudo-trimap $\mathcal{T}_{t,i}$, and the upsampled matte $\mathcal{A}_{t,i-1}$ from the preceding scale (for $i \ge 2$):
\begin{equation}
X_{t,i} =
\begin{cases}
\;\left[ F_{t,1}, \mathcal{T}_{t,1}, I_{t,1} \right], & i = 1, \\[4pt]
\;\left[ F_{t,i}, \mathcal{T}_{t,i}, I_{t,i}, U\!\left(\mathcal{A}_{\,t, i-1}\right) \right], & i = 2,\dots,n.
\end{cases}
\end{equation}
A scale-specific projection layer \(g_{\mathcal{A},i}(\cdot)\) first maps \(X_{t,i}\) to a fixed-dimensional embedding, from which a matting head \(f_{\mathcal{A},i}(\cdot)\) predicts the matte \(\mathcal{A}_{t,i}\) at scale \(i\):
\begin{equation}
\mathcal{A}_{t,i} =
\sigma\!\left(
f_{\mathcal{A},i}\!\left(
g_{\mathcal{A},i}(X_{t,i})
\right)
\right),
\quad
\mathcal{A}_{t,i} \in (0,1)^{H_i \times W_i}.
\end{equation}
Finally, the matte $\mathcal{A}_{t,n}$ at the finest scale $i = n$ is upsampled to the original resolution, yielding the final alpha matte $ \mathcal{A}_{t} \in (0,1)^{H \times W}$ for frame $ t $.

\subsection{Optimization Strategies}
\label{sec:optimization_strategies}

\textbf{Training Designs. } We freeze the VOS tracker to preserve its high-level tracking capability for superior temporal consistency, while training only the matting components on high-quality image matting data, enabling fine-grained alpha refinement without compromising the tracker's tracking robustness.

\noindent \textbf{Loss Designs. } Since \methodname is trained solely on images, supervision is applied per frame. For a training sample at index \(t\), we first threshold the ground-truth alpha matte \(\mathcal{A}^{\mathrm{GT}}_t\) within the range \([\alpha, \beta]\), and apply dilation and erosion to obtain the ground-truth ROI \(\mathcal{R}^{\mathrm{GT}}_t\).
\begin{equation}
\delta_t = \mathbf{1}\{\alpha \le \mathcal{A}^{\mathrm{GT}}_t \le \beta\}, \quad
\mathcal{R}^{\mathrm{GT}}_t = \mathrm{Dilate}(\delta_t) - \mathrm{Erode}(\delta_t).
\end{equation}
The ROI Detector is then optimized with a focal loss \( \mathcal{L}_{\mathrm{focal}} \) for pixel-wise binary classification and a smooth-\(L_1\) loss \( \mathcal{L}_{\mathrm{sm}} \) to reduce jagged artifacts \citep{ke2022modnet, wang2023video, yang2025matanyone}:
\begin{equation}
\mathcal{L}_{\mathcal{R}} =
\mathcal{L}_{\mathrm{focal}}(L_t, \mathcal{R}^{\mathrm{GT}}_t)
+
\mathcal{L}_{\mathrm{sm}}(L_t, \mathcal{R}^{\mathrm{GT}}_t).
\end{equation}
Following \citep{hou2019context, lin2022robust, li2024matting}, we adopt \(L_1\) loss \(\mathcal{L}_{L_1}\) and Laplacian loss \(\mathcal{L}_{\mathrm{lap}}\) for fine-grained alpha estimation. Inspired by the auxiliary loss in \citep{cheng2022masked}, we apply deep supervision across all prediction scales \(i \in \{1, \dots, n\}\), where \(\lambda_i\) is the loss weight for scale \(i\), and \(\mathcal{A}^{\mathrm{GT}}_{t,i}\) denotes the ground-truth alpha matte resized to scale \(i\):
\begin{equation}
\mathcal{L}_{\mathrm{alpha}} =
\sum_{i=1}^{n}
\lambda_i
\left(
\mathcal{L}_{L_1}(\mathcal{A}_{t,i}, \mathcal{A}^{\mathrm{GT}}_{t,i})
+
\mathcal{L}_{\mathrm{lap}}(\mathcal{A}_{t,i}, \mathcal{A}^{\mathrm{GT}}_{t,i})
\right).
\end{equation}
To preserve matte integrity and prevent hollow regions inside the matte foreground \citep{hou2019context, liu2021tripartite, li2023videomatt, li2024vmformer, huynh2024maggie}, we introduce a matte-mask consistency penalty \(\mathcal{L}_{\mathrm{con}}\) that anchors \(\mathcal{A}_t\) to the mask \(M_t\):
\begin{equation}
\mathcal{L}_{\mathrm{con}} = \mathcal{L}_{\mathrm{seg}}\bigl(\mathcal{A}_{t},\,M_{t}\bigr),
\end{equation}
where \(\mathcal{L}_{\mathrm{seg}}\) is a joint segmentation loss comprising focal and dice terms. The joint supervision for Progressive Alpha Predictor is then given by:
\begin{equation}
\mathcal{L}_{\mathcal{A}} =
\mathcal{L}_{\mathrm{alpha}}
+
\mathcal{L}_{\mathrm{con}}.
\end{equation}
Finally, the overall training objective $\mathcal{L}$ combining ROI detection and alpha refinement is formulated as:
\begin{equation}
\mathcal{L} =
\mathcal{L}_{\mathcal{R}}
+
\mathcal{L}_{\mathcal{A}}.
\end{equation}

\section{Experiments}

\subsection{Experimental Setup}
\label{sec:setup}

\textbf{Training Data.} We use 8 image matting datasets: I-HIM50K \citep{huynh2024maggie}, P3M-10k \citep{li2021privacy}, CelebAHairMask-HQ \citep{celebahairmaskhq}, AIM-500 \citep{li2021deep}, Distinctions-646 \citep{qiao2020attention}, AM-2K \citep{li2022bridging}, UHRIM \citep{yang2022exploring}, and RefMatte \citep{li2023referring}. For fair comparisons, we also evaluate variants with training datasets and trackers aligned to the baselines (Section~\ref{sec:ablation} and Table~\ref{tab:data_arch_ablation}).

\noindent \textbf{Implementation Details.} We develop three variants of \methodname using SAM2.1-Tiny, SAM2.1-Base+ \citep{ravi2024sam}, and SAM3 \citep{carion2025sam} as VOS trackers. Following our decoupled design, the tracker is kept frozen, with only the matting components optimized. All variants are trained for 5 epochs on 4 NVIDIA A6000 GPUs with a batch size of 32 using AdamW optimizer \citep{loshchilov2017decoupled}, with variant-specific learning rates. Hyperparameters are selected by grid search and set to $\theta = 0.65$, $\alpha = 0.15$, and $\beta = 0.5$. By default, the alpha predictor uses three scales, with loss weights \(\lambda_1 = 0.3\), \(\lambda_2 = 0.6\), and \(\lambda_3 = 1.2\).

\noindent \textbf{Metrics.} Following prior works \citep{xu2017deep, yao2024matte, huynh2024maggie, yang2025matanyone}, we adopt standard matting metrics: Mean Absolute Difference (MAD), Mean Squared Error (MSE), Gradient (Grad), Connectivity (Conn), and dtSSD (video matting only). For all metrics, lower values are better.

\subsection{Quantitative Evaluation on Image Matting}
\label{sec:image matting evaluation}
We evaluate the effectiveness of image matting on three benchmarks: P3M-500-NP \citep{li2021privacy}, AM-2K \citep{li2022bridging}, and PPM-100 \citep{ke2022modnet}. As shown in Table \ref{tab:img_matting_full}, all three variants of \methodname consistently outperform previous baselines across different metrics. For instance, the SAM2.1-Tiny variant achieves an 11.48 lower MAD than MAM on P3M-500-NP. Beyond the main comparison, we will further show in Section~\ref{sec:ablation} and Table~\ref{tab:data_arch_ablation} that the performance gains are mainly driven by the proposed matting design rather than merely by larger training data or a stronger tracker backbone.

\begin{table*}[t]
  \centering
  \captionsetup{justification=centering, skip=0.4pt}
    \caption{Quantitative results on image matting benchmarks. ``--'' denotes no reported result. The best, second-best, and third-best results are marked in \colorbox{rred}{\vphantom{dg}red}, \colorbox{oorange}{\vphantom{dg}orange}, and \colorbox{yyellow}{\vphantom{dg}yellow}, respectively}
    \vspace{1pt}
  \label{tab:img_matting_full}
  \scriptsize
  \renewcommand{\arraystretch}{1.1}
  \setlength{\tabcolsep}{3pt}
  \resizebox{\linewidth}{!}{
  \renewcommand{\arraystretch}{1.2}
  \begin{tabular}{lccccc@{\quad}ccccc@{\quad}ccccc}
    \toprule
    & \multicolumn{5}{c}{\textbf{P3M-500-NP}}
    & \multicolumn{5}{c}{\textbf{AM-2K test}}
    & \multicolumn{5}{c}{\textbf{PPM-100}} \\
    \cmidrule(r){2-6}\cmidrule(lr){7-11}\cmidrule(l){12-16}
    Methods
      & MAD$\downarrow$ & MSE$\downarrow$ & Grad$\downarrow$ & Conn$\downarrow$ & SAD$\downarrow$
      & MAD$\downarrow$ & MSE$\downarrow$ & Grad$\downarrow$ & Conn$\downarrow$ & SAD$\downarrow$
      & MAD$\downarrow$ & MSE$\downarrow$ & Grad$\downarrow$ & Conn$\downarrow$ & SAD$\downarrow$ \\
    \midrule
    P3M \citep{li2021privacy}
      & \ \ 6.50 & \ \ 3.50 & -- & -- & 11.23
      & 13.51 & \ \ 9.83 & 16.15 & 23.23 & 23.75
      & \ \ 9.60 & 5.80 & -- & 96.10 & 93.30 \\
    GFM \citep{li2022bridging}
      & -- & \ \ 5.60 & 14.80 & 18.00 & 15.50
      & \ \ 5.90 & \ \ 2.40 & \ \ 9.00 & \ \ 9.40 & \ \ 9.70
      & -- & -- & -- & -- & -- \\
    MODNet \citep{ke2022modnet}
      & 13.77 & \ \ 7.37 & 16.05 & 20.09 & 23.77
      & 36.28 & 27.39 & 17.38 & 59.49 & 62.77
      & \ \ 8.60 & 4.40 & 64.26 & 80.16 & 94.78 \\
    E2E-HIM \citep{liu2023end}
      & \ \ 5.40 & \ \ 3.00 & -- & -- & \ \ 9.25
      & -- & -- & -- & -- & --
      & \ \ 7.20 & 4.00 & -- & -- & -- \\
    MAM \citep{li2024matting}
      & 15.40 & \ \ 9.20 & 14.22 & -- & 25.82
      & 10.10 & \ \ 3.50 & 10.65 & -- & 17.30
      & \ \ 9.90 & 4.60 & 62.12 & 99.00 & 117.16 \\
    Lightweight \citep{zhong2024lightweight}
      & -- & -- & 10.78 & \ \ 9.77 & 10.60
      & -- & -- & -- & -- & --
      & -- & -- & 50.69 & 84.09 & 90.28 \\
    Matte Anything \citep{yao2024matte}
      & -- & \ \ 2.80 & 17.30 & 10.00 & 10.70
      & -- & \ \ 3.30 & 11.70 & 10.90 & 11.90
      & -- & -- & -- & -- & -- \\
    \textbf{\methodname (SAM2.1-T)}
      & \ \ \colorbox{yyellow}{3.92} & \ \ \colorbox{yyellow}{1.07} & \ \ \colorbox{oorange}{8.66} & \ \ \colorbox{yyellow}{6.34} & \ \ \colorbox{yyellow}{6.78}
      & \colorbox{oorange}{4.57} & \ \ \colorbox{oorange}{1.39} & \ \ \colorbox{oorange}{7.02} & \ \ \colorbox{oorange}{7.22} & \ \ \colorbox{oorange}{7.88}
      & \ \ \colorbox{yyellow}{4.51} & \colorbox{yyellow}{1.32} & \colorbox{yyellow}{49.26} & \colorbox{yyellow}{39.56} & \colorbox{yyellow}{42.05} \\
    \textbf{\methodname (SAM2.1-B+)}
      & \ \ \colorbox{rred}{3.81} & \ \ \colorbox{oorange}{1.00} & \ \ \colorbox{yyellow}{8.78} & \ \ \colorbox{oorange}{5.97} & \ \ \colorbox{rred}{6.58}
      & \colorbox{yyellow}{4.90} & \ \ \colorbox{yyellow}{1.59} & \ \ \colorbox{yyellow}{7.23} & \ \ \colorbox{yyellow}{7.75} & \ \ \colorbox{yyellow}{8.44}
      & \ \ \colorbox{oorange}{4.31} & \colorbox{oorange}{1.22} & \colorbox{oorange}{49.17} & \colorbox{oorange}{37.58} & \colorbox{oorange}{40.27} \\
    \textbf{\methodname (SAM3)}
      & \ \ \colorbox{oorange}{3.83} & \ \ \colorbox{rred}{0.97} & \ \ \colorbox{rred}{8.48} & \ \ \colorbox{rred}{5.84} & \ \ \colorbox{oorange}{6.61}
      & \colorbox{rred}{4.32} & \ \ \colorbox{rred}{1.19} & \ \ \colorbox{rred}{6.75} & \ \ \colorbox{rred}{6.61} & \ \ \colorbox{rred}{7.43}
      & \ \ \colorbox{rred}{4.23} & \colorbox{rred}{1.16} & \colorbox{rred}{47.55} & \colorbox{rred}{36.27} & \colorbox{rred}{39.45} \\
    \bottomrule
  \end{tabular}
  }
\end{table*}

\begin{table*}[t]
  \centering
  \captionsetup{justification=centering, skip=0.4pt}
  \caption{Quantitative results on video matting benchmarks. \methodname is evaluated \textbf{zero-shot}.}
  \label{tab:video_matting_full}
  \scriptsize
  \renewcommand{\arraystretch}{1.1}
  \setlength{\tabcolsep}{2pt}
  \resizebox{\textwidth}{!}{
  \begin{tabular}{lc*{15}{c}}
    \toprule
    & & \multicolumn{5}{c}{\textbf{V-HIM60-Medium}}
      & \multicolumn{5}{c}{\textbf{V-HIM60-Hard}}
      & \multicolumn{5}{c}{\textbf{VideoMatte-SD}} \\
    \cmidrule(r){3-7}
    \cmidrule(lr){8-12}
    \cmidrule(l){13-17}
    Methods & Venue
      & MAD$\downarrow$ & MSE$\downarrow$ & Grad$\downarrow$ & Conn$\downarrow$ & dtSSD$\downarrow$
      & MAD$\downarrow$ & MSE$\downarrow$ & Grad$\downarrow$ & Conn$\downarrow$ & dtSSD$\downarrow$
      & MAD$\downarrow$ & MSE$\downarrow$ & Grad$\downarrow$ & Conn$\downarrow$ & dtSSD$\downarrow$ \\
    \midrule
    RVM \citep{lin2022robust} & \pub{WACV'22}
      & -- & -- & -- & -- & --
      & -- & -- & -- & -- & --
      & \ \ 6.08 & \ \ 1.47 & \ \ 0.88 & \ \ 0.41 & \ \ 1.36 \\
    InstMatt \citep{sun2022human} & \pub{CVPR'22}
      & 19.34 & -- & \ \ 7.21 & \ \ 6.02 & 24.98
      & \ \ 27.24 & -- & \ \ 7.88 & \ \ 8.02 & 31.89
      & -- & -- & -- & -- & -- 
      \\
    FTP-VM \citep{huang2023end} & \pub{CVPR'23}
      & 26.86 & -- & 12.39 & \ \ 9.95 & 32.64
      & \ \ 48.11 & -- & 14.87 & 16.12 & 45.29
      & \ \ 6.13 & \ \ 1.31 & \ \ 1.14 & \ \ 0.41 & \ \ 1.60 \\
    AdaM \citep{lin2023adaptive} & \pub{CVPR'23}
      & -- & -- & -- & -- & --
      & -- & -- & -- & -- & --
      & \ \ 5.30 & \ \ 0.78 & \ \ 0.72 & \ \ 0.30 & \ \ 1.33 
      \\
    SparseMat \citep{sun2023ultrahigh} & \pub{CVPR'23}
      & 18.20 & -- & \ \ 8.03 & \ \ 6.87 & 30.19
      & \ \ 24.83 & -- & \ \ 8.47 & \ \ 8.19 & 36.92
      & -- & -- & -- & -- & -- 
      \\
    MaGGIe \citep{huynh2024maggie} & \pub{CVPR'24}
      & 13.85 & -- & \ \ \colorbox{oorange}{6.31} & \ \ 5.11 & 23.63
      & \ \ 21.23 & -- & \ \ \colorbox{oorange}{7.08} & \ \ 6.89 & 29.90
      & \ \ 5.49 & \ \ 0.60 & \ \ 0.57 & \ \ 0.31 & \ \ 1.39 \\
    MatAnyone \citep{yang2025matanyone} & \pub{CVPR'25}
      & 29.95 & 19.72 & \ \ 9.03 & 12.28 & \ \ 5.98
      & \ \ 30.09 & 18.87 & \ \ 8.93 & 10.00 & \ \ 6.72
      & \ \ 5.15 & \ \ 0.93 & \ \ 0.67 & \ \ 0.26 & \ \ 1.18 \\
    MatAnyone2 \citep{yang2025matanyone2} & \pub{CVPR'26}
      & 15.12 & \ \ 5.86 & \ \ \colorbox{yyellow}{6.36} & \ \ 5.43 & \ \ 4.50
      & \ \ 45.75 & 35.03 & \ \ 8.43 & 14.75 & \ \ 6.16
      & \ \ \colorbox{oorange}{4.73} & \ \ 0.55 & \ \ 0.51 & \ \ \colorbox{oorange}{0.19} & \ \ \colorbox{oorange}{1.12} 
      \\
    \textbf{\methodname (SAM2.1-T)} & --
      & \colorbox{yyellow}{13.76} & \ \ \colorbox{oorange}{4.61} & \ \ 7.78 & \ \ \colorbox{yyellow}{5.01} & \ \ \colorbox{oorange}{4.23}
      & \ \ \colorbox{yyellow}{18.58} & \ \ \colorbox{yyellow}{8.79} & \ \ 8.03 & \ \ \colorbox{yyellow}{6.16} & \ \ \colorbox{yyellow}{5.37}
      & \ \ 4.85 & \ \ \colorbox{yyellow}{0.41} & \ \ \colorbox{yyellow}{0.36} & \ \ \colorbox{yyellow}{0.20} & \ \ 1.22 
      \\
    \textbf{\methodname (SAM2.1-B+)} & --
      & \colorbox{oorange}{13.71} & \ \ \colorbox{yyellow}{4.74} & \ \ 7.28 & \ \ \colorbox{oorange}{4.89} & \ \ \colorbox{yyellow}{4.24}
      & \ \ \colorbox{oorange}{18.20} & \ \ \colorbox{oorange}{8.55} & \ \ \colorbox{yyellow}{7.39} & \ \ \colorbox{oorange}{6.01} & \ \ \colorbox{oorange}{5.10}
      & \ \ \colorbox{yyellow}{4.83} & \ \ \colorbox{oorange}{0.36} & \ \ \colorbox{oorange}{0.34} & \ \ \colorbox{oorange}{0.19} & \ \ \colorbox{yyellow}{1.15} \\
    \textbf{\methodname (SAM3)} & --
      & \colorbox{rred}{11.77} & \ \ \colorbox{rred}{3.64} & \ \ \colorbox{rred}{5.92} & \ \ \colorbox{rred}{4.23} & \ \ \colorbox{rred}{3.81}
      & \ \ \colorbox{rred}{14.37} & \ \ \colorbox{rred}{5.52} & \ \ \colorbox{rred}{5.85} & \ \ \colorbox{rred}{4.72} & \ \ \colorbox{rred}{4.37}
      & \ \ \colorbox{rred}{4.44} & \ \ \colorbox{rred}{0.27} & \ \ \colorbox{rred}{0.23} & \ \ \colorbox{rred}{0.16} & \ \ \colorbox{rred}{1.11} \\
    \bottomrule
  \end{tabular}
  }
\end{table*}

\subsection{Quantitative Evaluation on Video Matting}
\label{sec:video matting evaluation}

We benchmark \methodname's video matting performance on V-HIM60~\citep{huynh2024maggie} and VideoMatte~\citep{lin2021real} in a \textit{zero-shot} manner, comparing against baselines supervised on video matting datasets. These baselines include recent SOTA approaches such as MatAnyone2 \citep{yang2025matanyone2}, MatAnyone \citep{yang2025matanyone}, MaGGIe \citep{huynh2024maggie}, FTP-VM \citep{huang2023end}, as well as RVM \citep{lin2022robust}. As shown in Table~\ref{tab:video_matting_full}, \methodname consistently outperforms these video-supervised baselines despite its zero-shot setting. It also achieves the lowest dtSSD, indicating strong temporal consistency inherited from the VOS tracker. These results validate our decoupled design, which enables tracking and matting to specialize independently yet cooperate effectively for robust video matting.

\subsection{Qualitative Evaluation}

\begin{figure*}[!t]
  \centering
  \includegraphics[width=1.0\textwidth]{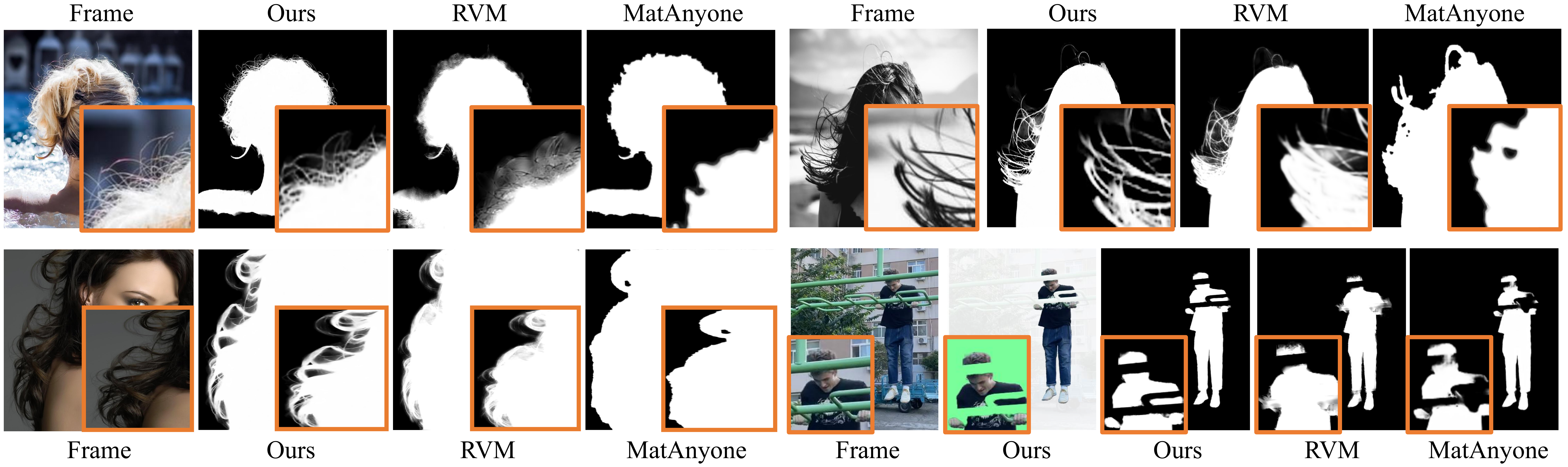}
  \vspace{-0.3in}
  \caption{Qualitative comparison with MatAnyone \citep{yang2025matanyone} and RVM \citep{lin2022robust} on human matting. \methodname demonstrates superior capabilities in resolving fine-grained details of hair strands and semi-transparencies. (Zoom in for details)}
  \label{fig:figure4}
\end{figure*}

\noindent \textbf{Human Matting.} \methodname outperforms competitive baselines including RVM \citep{lin2022robust} and MatAnyone \citep{yang2025matanyone}, producing superior results on intricate hair-level details and semi-transparencies. As shown in Figure \ref{fig:figure4}, \methodname preserves fine details under complex lighting while suppressing undesired occlusion parts of the foreground, such as the green bars in front of the human.

\begin{figure*}[!t]
  \centering
  \includegraphics[width=1.0\textwidth]{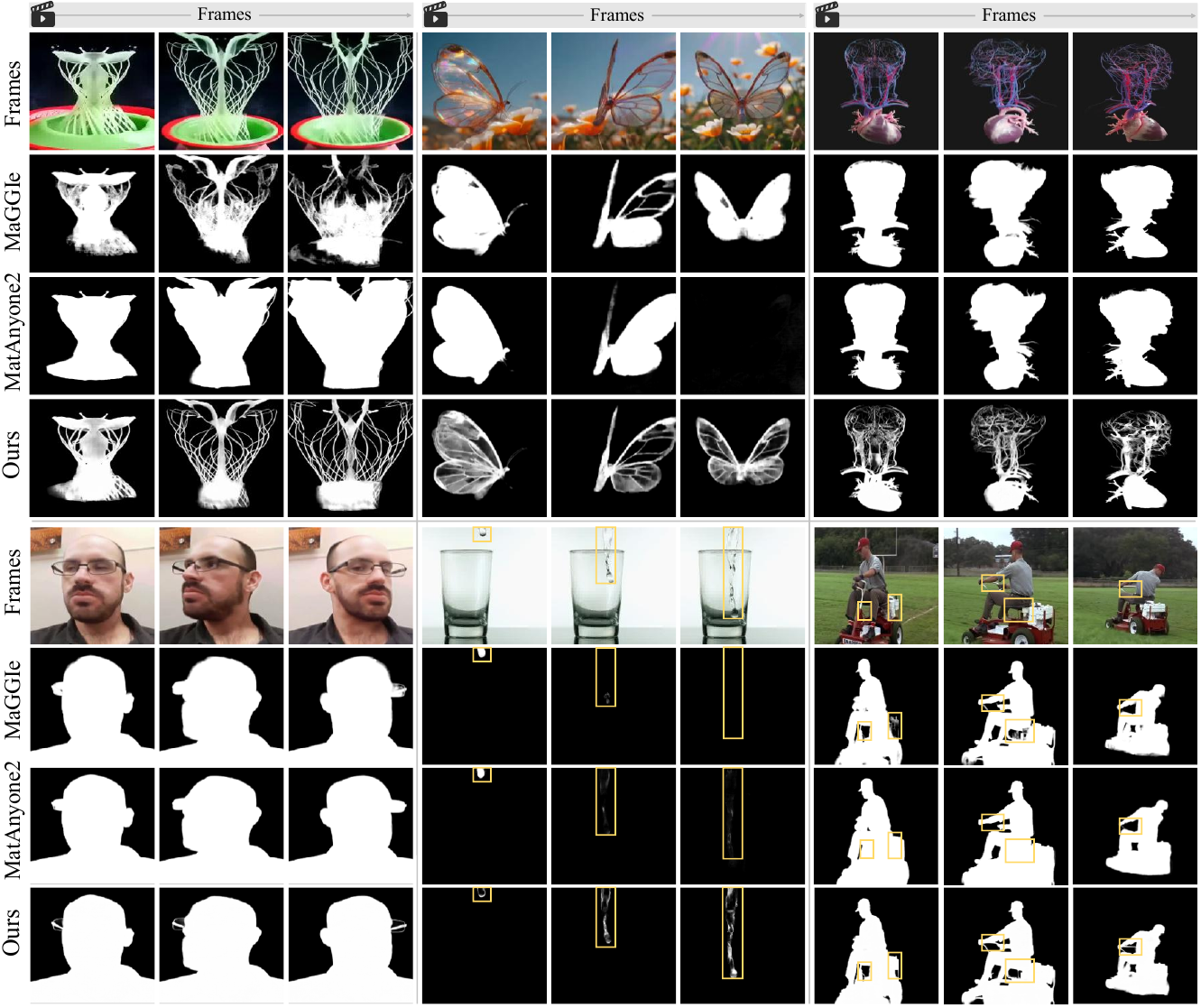}
  \vspace{-0.28in}
  \caption{Qualitative comparison with MatAnyone2 \citep{yang2025matanyone2} and MaGGIe \citep{huynh2024maggie} on in-the-wild sequences. The baselines struggle with non-human targets and rapid motion, while \methodname preserves stable tracking and recovers intricate details. (Zoom in for details)}
  \label{fig:figure5_1}
  \vspace{-5mm}
\end{figure*}

\noindent \textbf{In-the-wild Matting.} As shown in Figure~\ref{fig:figure5_1}, existing SOTA video matting methods (MatAnyone2 \citep{yang2025matanyone2}, MaGGIe \citep{huynh2024maggie}) trained on domain-specific, often human-centric video matting datasets struggle to generalize to in-the-wild sequences, especially fast-moving targets such as the rapidly growing roots, semi-transparent butterflies, and rapid-dripping water. In contrast, our decoupled strategy preserves robust high-level tracking from which dedicated matting components reliably extract fine details.

\begin{figure*}[!t]
  \centering
  \includegraphics[width=1.0\textwidth]{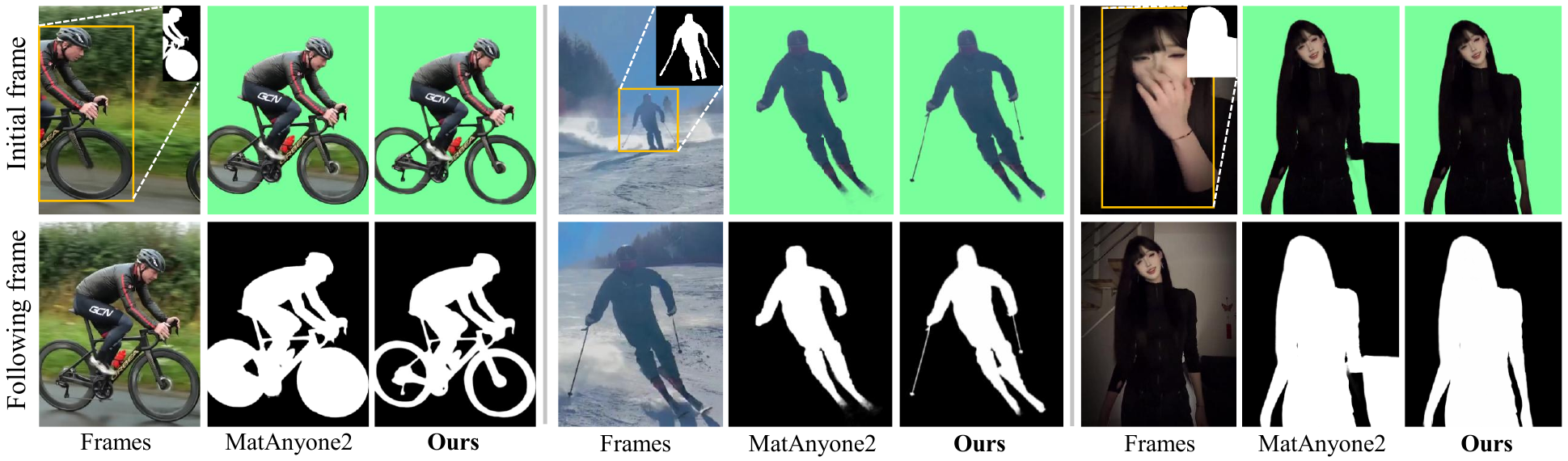}
  \vspace{-0.28in}
  \caption{\methodname robustly preserves target attachments (e.g., bicycles and ski poles) while suppressing closely-attached background distractions. (Zoom in for details)}
  \label{fig:figure5_2}
\end{figure*}

\noindent \textbf{Target Attachments and Background Distractions.} As shown in Figure~\ref{fig:figure5_2}, \methodname effectively handles targets with attached objects, such as people riding bicycles (left) or holding ski poles (middle). It also suppresses background distractions, such as the desk right to the woman (right), benefiting from the matte-mask consistency supervision outlined in Section~\ref{sec:optimization_strategies}.

\subsection{Ablation Studies}
\label{sec:ablation}
We employ \methodname (SAM2.1-B+) as our default model to conduct the ablation studies.

\noindent \textbf{Controlled Comparison with Baselines.}
We compare \methodname with MAM \citep{li2024matting} and Matte Anything \citep{yao2024matte} under controlled settings to show that our performance gain does not rely on larger training data or a stronger backbone. Specifically, we train \methodname on the same datasets used by each baseline (Table \ref{tab:data_arch_ablation}, left), and then equip both baselines with the same SAM2.1-B+ tracker (Table \ref{tab:data_arch_ablation}, right). As shown in Table~\ref{tab:data_arch_ablation}, \methodname performs consistently better in both scenarios, indicating that the benefit is driven more by its architectural and supervision designs. We next ablate these designs in detail.

\begin{figure*}[!t]
  \centering
  \includegraphics[width=1.0\textwidth]{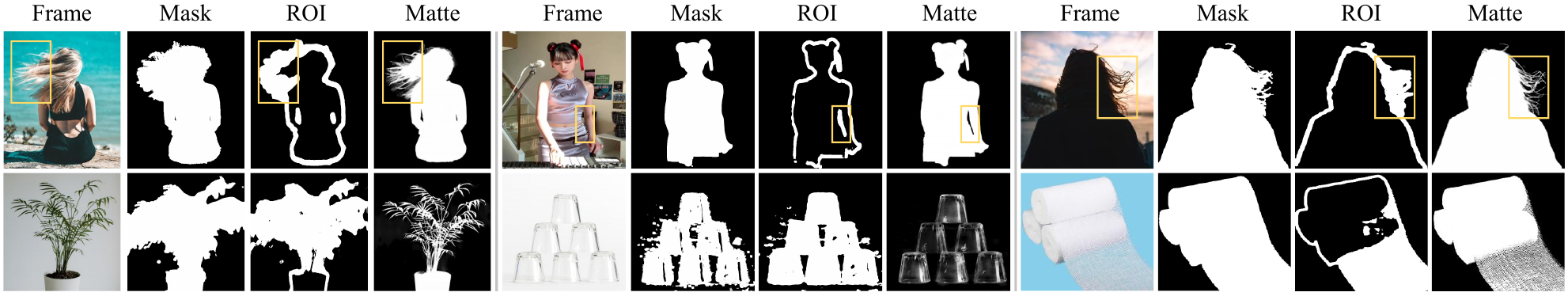}
  \vspace{-0.2in}
  \caption{The ROI Detector identifies instance-specific matting-critical regions missed by morphological operations or raw masks, such as flying hair, thin leaves, and limb gaps. (Zoom in for details)}
  \label{fig:ablation_roi_new}
\end{figure*}

\noindent \textbf{ROI Strategies.}
Table \ref{tab:ablation_ROI} ablates different ROI strategies on V-HIM60-Hard \citep{huynh2024maggie}. We compare \methodname against two baselines: (a) ``Morphological'', which generates ROI using standard dilation and erosion on the mask \citep{sharma2020alphanet, zhou2021semantic, yao2024matte}; and (b) ``Mask-only'', which directly uses the raw VOS mask for matte prediction \citep{yu2021mask, park2023mask, li2024matting, yang2025matanyone}. Our ROI Detector outperforms both baselines across all metrics, highlighting its ability to identify subtle regions with fine details and semi-transparency. Figure \ref{fig:ablation_roi_new} visualizes this effect, where the ROI Detector captures instance-specific matting-critical regions, such as the flying hair, intricate leaves and arm gaps easily overlooked by morphological operations or raw masks.

\begin{table*}[t]
  \centering
  \captionsetup{justification=centering, skip=1.5pt}
  \caption{Controlled comparison with baseline methods. Left: aligned training data. Right: aligned tracker backbone.}
  \label{tab:data_arch_ablation}
  \scriptsize
  \renewcommand{\arraystretch}{1.1}
  \setlength{\tabcolsep}{3pt}
  \begin{minipage}{0.49\textwidth}
    \centering
    \textbf{(a) Aligned Training Data (* / \dag)}\\[2pt]
    \resizebox{\linewidth}{!}{
    \begin{tabular}{lc ccc ccc}
      \toprule
      & & \multicolumn{3}{c}{\textbf{P3M-500-NP}}
      & \multicolumn{3}{c}{\textbf{AM-2K test}} \\
      \cmidrule(r){3-5}\cmidrule(l){6-8}
      Methods & Data
        & MAD$\downarrow$ & MSE$\downarrow$ & Grad$\downarrow$
        & MAD$\downarrow$ & MSE$\downarrow$ & Grad$\downarrow$ \\
      \midrule
      MAM 
        & *
        & 15.40 & \ \ 9.20 & 14.22
        & 10.10 & \ \ 3.50 & 10.65 \\
      Matte Anything 
        & \dag
        & -- & \ \ 2.80 & 17.30
        & -- & \ \ 3.30 & 11.70 \\
        \rowcolor{defaultColor} \textbf{\methodname}
        & *
        & \ \ \textbf{4.05} & \ \ \textbf{1.14} & \ \ \textbf{9.58}
        & \ \ \textbf{5.38} & \ \ \textbf{1.91} & \ \ \textbf{7.81} \\
      \rowcolor{defaultColor} \textbf{\methodname}
        & \dag
        & \ \ \textbf{3.94} & \ \ \textbf{1.11} & \ \ \textbf{8.85}
        & \ \ \textbf{5.95} & \ \ \textbf{2.48} & \ \ \textbf{8.25} \\
      \bottomrule
    \end{tabular}
    }
  \end{minipage}
  \hfill
    \begin{minipage}{0.49\textwidth}
    \centering
    \textbf{(b) Aligned Tracker Backbone (SAM2.1-B+)}\\[2pt]
    \resizebox{\linewidth}{!}{
    \renewcommand{\arraystretch}{1.5}
    \begin{tabular}{lccc ccc}
      \toprule
      & \multicolumn{3}{c}{\textbf{P3M-500-NP}}
      & \multicolumn{3}{c}{\textbf{AM-2K test}} \\
      \cmidrule(r){2-4}\cmidrule(l){5-7}
      Methods
        & MAD$\downarrow$ & MSE$\downarrow$ & Grad$\downarrow$
        & MAD$\downarrow$ & MSE$\downarrow$ & Grad$\downarrow$ \\
      \midrule
      MAM (SAM2.1-B+)
        & 12.92 & \ \ 6.08 & 11.64
        & \ \ 8.64 & \ \ 2.34 & \ \ 8.78 \\
      Matte Anything (SAM2.1-B+)
        & \ \ 6.00 & \ \ 1.61 & 13.20
        & \ \ 6.21 & \ \ 1.67 & \ \ 8.67 \\
      \rowcolor{defaultColor} \textbf{\methodname (SAM2.1-B+)}
        & \ \ \textbf{3.81} & \ \ \textbf{1.00} & \ \textbf{8.78}
        & \ \ \textbf{4.90} & \ \ \textbf{1.59} & \ \ \textbf{7.23} \\
      \bottomrule
    \end{tabular}
    }
    \end{minipage}
\end{table*}

\begin{table*}[!ht]
  \centering
  \scriptsize
  \begin{minipage}[t]{0.44\linewidth}
    \centering
    \captionsetup{justification=centering, skip=0.5pt}
    \caption{Ablation on different ROI strategies.}
    \label{tab:ablation_ROI}
    \renewcommand{\arraystretch}{1.48} 
    \setlength{\tabcolsep}{5.5pt}
    \begin{tabular}{l l ccccc} 
      \toprule
      & \makecell[c]{ROI \\ Strategies} & \makecell[c]{MAD \\ $\downarrow$}  & \makecell[c]{Grad \\ $\downarrow$}  & \makecell[c]{Conn \\ $\downarrow$}  & \makecell[c]{dtSSD \\ $\downarrow$} \\
      \midrule
      (a) & Morphological & 29.82 & 11.57 & 10.37 & 7.48 \\
      (b) & Mask-only & 20.07 & \ \ 9.11 & \ \ 6.68 & 5.50 \\
      \rowcolor{defaultColor} (c) & \textbf{ROI Detector} & \textbf{18.20} & \ \ \textbf{7.39} & \ \ \textbf{6.01} & \textbf{5.10} \\
      \bottomrule
    \end{tabular}
  \end{minipage}
  \hfill
  \begin{minipage}[t]{0.53\linewidth}
    \centering
    \captionsetup{justification=centering, skip=0.5pt}
    \caption{Ablation on architectural and supervision designs.}
    \label{tab:ablation_loss}
    \renewcommand{\arraystretch}{1.1}
    \setlength{\tabcolsep}{5.5pt}
    \begin{tabular}{l ccc cccc}
      \toprule
       & \makecell[c]{Prog.\\Scaling} & \makecell[c]{Con.\\ Loss} & \makecell[c]{Smooth\\Loss} & \makecell[c]{MAD \\ $\downarrow$} & \makecell[c]{Grad \\ $\downarrow$} & \makecell[c]{Conn \\ $\downarrow$} & \makecell[c]{dtSSD \\ $\downarrow$} \\
      \midrule
      (a) &  & \checkmark & \checkmark & 19.43 & 7.88 & 6.35 & 5.30 \\
      (b) & \checkmark & & \checkmark & 18.65 & 7.70 & 6.20 & 5.18 \\
      (c) & \checkmark & \checkmark &  & 18.26 & 7.45 & 6.04 & \textbf{5.09} \\
      \rowcolor{defaultColor} (d) & \checkmark & \checkmark & \checkmark & \textbf{18.20} & \textbf{7.39} & \textbf{6.01} & 5.10 \\
      \bottomrule
    \end{tabular}
  \end{minipage}
\end{table*}

\begin{table*}[!ht]
  \centering
  \captionsetup{skip=0.5pt}
  \caption{Ablation of fine-tuning on large-scale video matting dataset. ``Original'' denotes the original decoupled model, while ``Video-FT'' denotes its fine-tuned variant on V-HIM2K5 \citep{huynh2024maggie}.}
  \label{tab:ablation_video}
  \scriptsize
  \renewcommand{\arraystretch}{1.1}
  \begin{tabular}{l ccc ccc ccc ccc}
    \toprule
    & \multicolumn{3}{c}{V-HIM60-Hard} & \multicolumn{3}{c}{VideoMatte-SD} & \multicolumn{3}{c}{AM-2K test} & \multicolumn{3}{c}{PPM-100} \\
    \cmidrule(lr){2-4} \cmidrule(lr){5-7} \cmidrule(lr){8-10} \cmidrule(lr){11-13}
    Methods & MAD$\downarrow$ & Grad$\downarrow$ & dtSSD$\downarrow$ & MAD$\downarrow$ & Grad$\downarrow$ & dtSSD$\downarrow$ & MAD$\downarrow$ & Grad$\downarrow$ & Conn$\downarrow$ & MAD$\downarrow$ & Grad$\downarrow$ & Conn$\downarrow$ \\
    \midrule
    Video-FT & \textbf{17.90} & \textbf{7.08} & \textbf{5.01} & 4.85 & \textbf{0.34} & \textbf{1.12} & 5.23 & 7.40 & 7.96 & 4.40 & 49.63 & 38.19 \\
    \rowcolor{defaultColor} \textbf{Original} & 18.20 & 7.39 & 5.10 & \textbf{4.83} & \textbf{0.34} & 1.15 & \textbf{4.90} & \textbf{7.23} & \textbf{7.75} & \textbf{4.31} & \textbf{49.17} & \textbf{37.58} \\
    \bottomrule
  \end{tabular}
\end{table*}

\begin{figure*}[!t]
  \centering
  \includegraphics[width=1.0\textwidth]{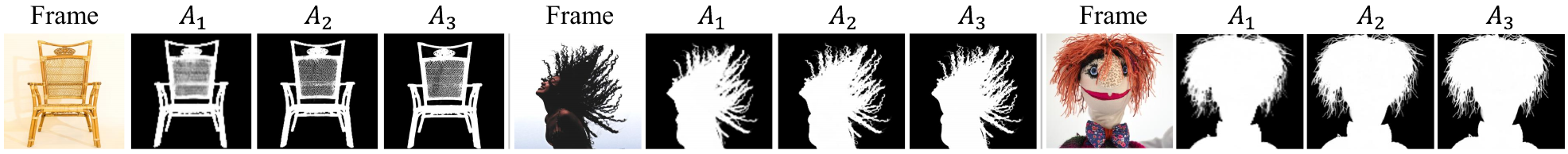}
  \vspace{-0.2in}
  \caption{The intermediate mattes are iteratively refined and improved across scales (from $\mathcal{A}_{1}$ to $\mathcal{A}_{3}$), progressively capturing finer details. (Zoom in for details)}
  \label{fig:ablation_three_new}
\end{figure*}

\begin{figure*}[!ht]
  \centering
  \includegraphics[width=1.0\textwidth]{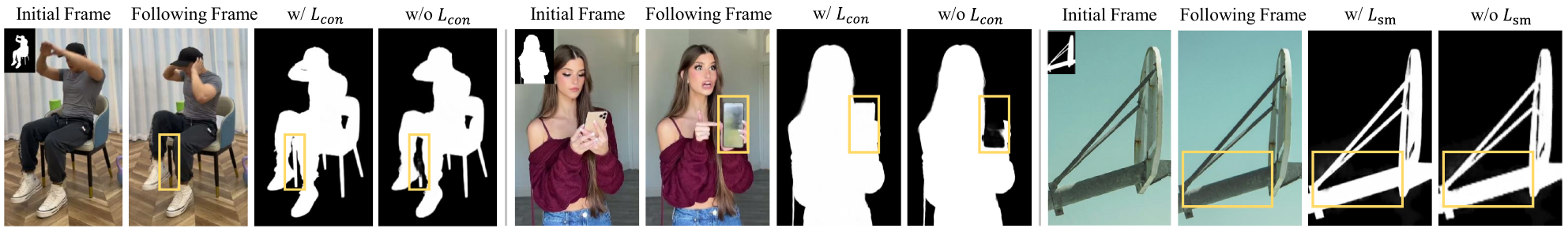}
  \vspace{-0.2in}
  \caption{\textit{Left and middle}: The matte-mask consistency loss $\mathcal{L}_{con}$ prevents hollow foreground regions and preserves matte integrity. \textit{Right}: The smoothness loss $\mathcal{L}_{sm}$ reduces jagged edges for cleaner mattes. (Zoom in for details)}
  \label{fig:ablation_vis_2}
\end{figure*}

\noindent \textbf{Architecture and Supervision Designs.}
Table \ref{tab:ablation_loss} (a) validates the multi-scale cascade refinement in the progressive alpha predictor. Figure \ref{fig:ablation_three_new} shows the predictor exhibiting a clear coarse-to-fine refinement across scales (from $\mathcal{A}_{1}$ to $\mathcal{A}_{3}$), progressively recovering fine structures such as the hollow chair and flying hair. Table \ref{tab:ablation_loss} (b,c) further confirms the effectiveness of our supervision designs: the matte-mask consistency penalty (b) fills foreground holes (Figure \ref{fig:ablation_vis_2}, left and middle), while the smoothness loss (c) reduces jagged boundaries (Figure \ref{fig:ablation_vis_2}, right). Together, these designs contribute to high-fidelity video matting.

\begin{figure*}[!ht]
  \centering
  \includegraphics[width=1.0\textwidth]{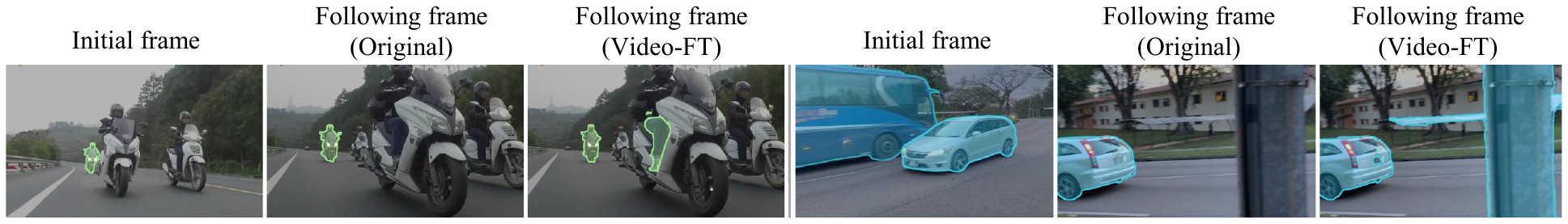}
  \vspace{-0.2in}
  \caption{Comparison of tracking robustness. ``Original'' denotes the original decoupled model, while ``Video-FT'' denotes the variant fine-tuned on video matting data. Video fine-tuning noticeably weakens the original tracking robustness. (Zoom in for details)}
  \label{fig:video-ft}
\end{figure*}

\noindent \textbf{Does fine-tuning on video matting datasets help?}
We investigate this by fine-tuning our model on V-HIM2K5 \citep{huynh2024maggie}, a public large-scale human-centric video matting dataset. Table \ref{tab:ablation_video} shows a clear trade-off: fine-tuning improves the in-domain benchmark (V-HIM60-Hard) but degrades generalization, with worse results on out-of-domain animal data (AM-2K) and little change on other human-centric datasets (VideoMatte and PPM-100). This indicates overfitting, as the video matting data covers relatively simple scenarios within a narrow domain. Figure \ref{fig:video-ft} further illustrates this effect, where the video-fine-tuned variant of \methodname noticeably degrades the original tracking robustness, even in simple scenarios without occlusions.

\subsection{Resistance to Tracking Inaccuracies}
\begin{figure*}[!t]
  \centering
  \includegraphics[width=1.0\textwidth]{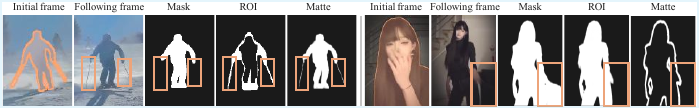}
  \vspace{-0.3in}
  \caption{\methodname rectifies tracking inaccuracies. \textit{Left}: The ROI recovers the ski poles missed by the VOS mask. \textit{Right}: The ROI removes the background desk mistakenly included by the VOS mask. (Zoom in for details)}
  \label{fig:rectify}
  \vspace{-0.2em}
\end{figure*}

\methodname can be robust to inaccuracies induced by its VOS tracker. Rather than using the VOS mask as a hard constraint, the ROI detector treats the mask as one of multiple cues and fuses it with image-level appearance priors for ROI detection, as illustrated in Figure \ref{fig:figure2}. This allows the predicted ROI to suppress tracker-induced errors and provide reliable guidance for the subsequent alpha predictor. As shown in Figure \ref{fig:rectify}, \methodname recovers missing foreground details overlooked by its tracker (e.g., ski poles, left) and removes wrongly included background objects (e.g., the desk, right), enabling precise video matting.

\subsection{FPS and VRAM Efficiency}
\label{sec:fps_compare}
\vspace{-0.05in}
We evaluate the computational efficiency of \methodname on the VideoMatte \citep{lin2021real} benchmark using a single NVIDIA A6000 GPU. As shown in Table~\ref{tab:efficiency_comparison}, all three variants maintain stable FPS and modest VRAM usage across different input resolutions. Notably, both the SAM2.1-T and SAM2.1-B+ variants process videos at over 30 FPS, enabling real-time video matting.

\subsection{Flexible Prompting.}
\vspace{-0.05in}
\methodname supports prompt types inherited from the VOS tracker, enabling interactive video matting. Beyond initial-frame masks, all variants of \methodname support points and boxes, while text prompts are supported in the SAM3 variant. Figure \ref{fig:prompting} shows a selfie-video case where different prompts all produce high-quality results. This flexibility makes video matting easier and more accessible for real-world applications.

\vspace{-0.5em}
\section{Discussion}
\begin{figure}[!t]
  \centering
  \begin{minipage}[t]{0.57\columnwidth}
    \centering
    \includegraphics[width=\linewidth]{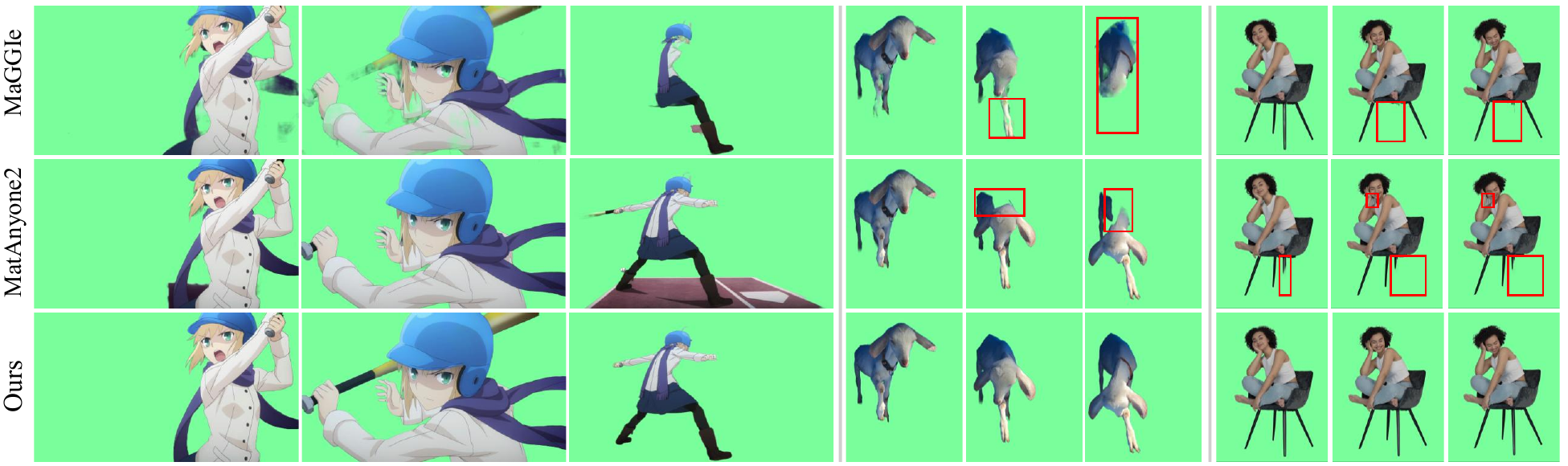}
    \vspace{-0.28in}
    \caption{Qualitative comparison of the flickering effect. (Zoom in for details)}
    \label{fig:flicker}
  \end{minipage}
  \hfill
  \begin{minipage}[t]{0.41\columnwidth}
    \centering
    \includegraphics[width=\linewidth,height=0.12\textheight]{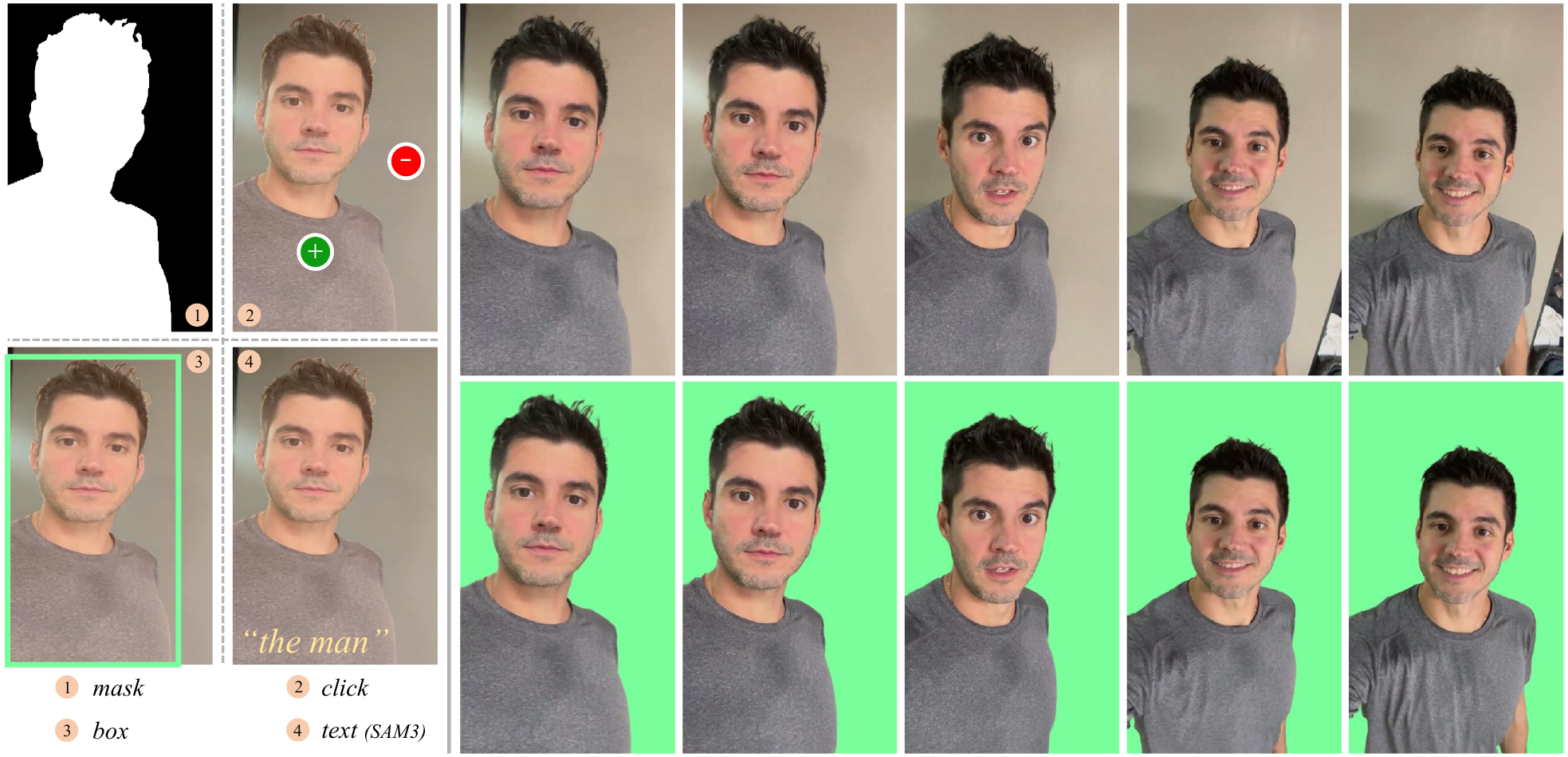}
    \vspace{-0.28in}
    \caption{Flexible prompting with different prompt types. (Zoom in for details)}
    \label{fig:prompting}
  \end{minipage}
  \vspace{-0.1em}
\end{figure}

\vspace{-0.05in}
\noindent \textbf{Flickering effect.}
Flickering reflects temporal inconsistency in video matting. Figure~\ref{fig:flicker} shows that \methodname maintains stable mattes under rapid target motion, preserving coherent foreground over time. In contrast, the baselines exhibit more pronounced temporal fluctuations and visible flickering artifacts.

\begin{figure}[!t]
  \centering
  \captionsetup{justification=centering}
  \includegraphics[width=1.0\columnwidth]{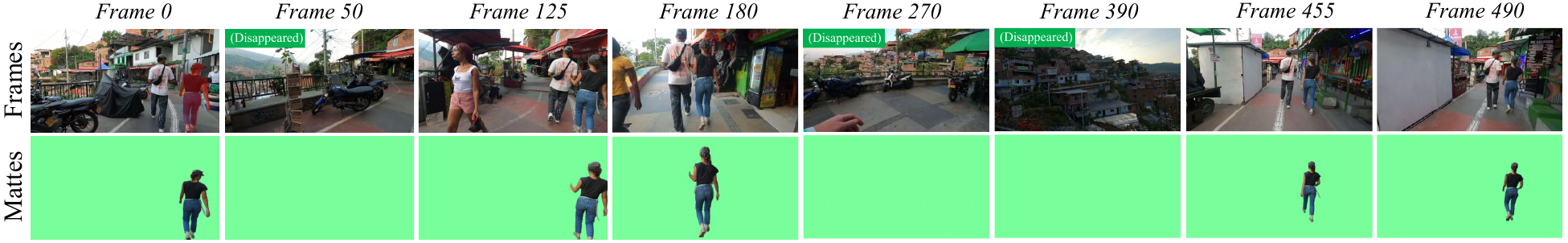}
  \vspace{-0.28in}
  \caption{Stable matting on a long video with frequent target occlusions and reappearances. (Zoom in for details)}
  \label{fig:long}
  \vspace{-0.2em}
\end{figure}

\begin{table*}[!t]
\centering
\caption{FPS and VRAM efficiency comparison on the VideoMatte benchmark, measured on a single NVIDIA A6000 GPU. ``oom'' denotes Out of Memory.}
\vspace{-0.3em}
\label{tab:efficiency_comparison}
\begin{minipage}[t]{0.48\textwidth}
\centering
\footnotesize
(a) FPS $\uparrow$ \\ \vspace{1mm}
\scriptsize
\renewcommand{\arraystretch}{1.1}
\setlength{\tabcolsep}{6pt}
\resizebox{\textwidth}{!}{
\begin{tabular}{lcccc}
\toprule
Methods & 720p & 1080p & 1440p & 2160p \\
\midrule
MatAnyone {\tiny \citep{yang2025matanyone}}    & 25.96 & 11.29 & \ 2.92 & oom \\
MatAnyone2 {\tiny \citep{yang2025matanyone2}}  & 21.94 & \ 9.93  & \ 2.82 & oom \\
\midrule
\textbf{\methodname (SAM2.1-T)}      & 40.46    & 40.31    & 40.21  & 40.04 \\
\textbf{\methodname (SAM2.1-B+)}     & 30.40    & 30.36    & 30.28  & 30.23 \\
\textbf{\methodname (SAM3)}         & \ 9.09    & \ 9.07    &  \ 9.02    & \ 8.99 \\
\bottomrule
\end{tabular}
}
\end{minipage}
\begin{minipage}[t]{0.48\textwidth}
\centering
\footnotesize
(b) VRAM (GB) $\downarrow$ \\ \vspace{1mm}
\scriptsize
\renewcommand{\arraystretch}{1.1}
\setlength{\tabcolsep}{6pt}
\resizebox{\textwidth}{!}{
\begin{tabular}{lcccc}
\toprule
Methods & 720p & 1080p & 1440p & 2160p \\
\midrule
MatAnyone {\tiny \citep{yang2025matanyone}}    & 3.63 & 14.16 & 41.31 & oom \\
MatAnyone2 {\tiny \citep{yang2025matanyone2}}  & 3.10  & 13.67 & 41.28 & oom  \\
\midrule
\textbf{\methodname (SAM2.1-T)}      & 3.08     & \ \ 3.61    & \ \ 4.71    & 6.25  \\
\textbf{\methodname (SAM2.1-B+)}     & 3.42     & \ \ 3.82    & \ \ 4.88    & 6.45  \\
\textbf{\methodname (SAM3)}         & 4.80     & \ \ 4.91    & \ \ 5.44    & 6.88  \\
\bottomrule
\end{tabular}
}
\end{minipage}
\end{table*}

\noindent \textbf{Performance on long videos.}
Practical applications, such as film post-production and e-commerce live streaming, require video matting to handle long sequences where the target may enter and exit the scene. However, existing public benchmarks mostly contain short clips. We therefore demonstrate \methodname on a challenging long video. As shown in Figure \ref{fig:long}, despite repeated disappearances and reappearances of the target, \methodname consistently tracks and mattes across 500 frames, demonstrating stable matting over extended sequences.

\section{Conclusion}

We present \methodname, a generalized image and video matting framework that decouples high-level tracking from low-level matting. \methodname achieves state-of-the-art video matting performance without relying on costly video matting datasets, while generalizing robustly to both human-centric and in-the-wild scenarios. It offers real-time efficiency, supports diverse prompt types, and maintains strong temporal consistency over extended videos. We expect \methodname to facilitate real-world deployment and inspire future research.

\bibliographystyle{plainnat}
\bibliography{mybib.bib}

@String(IJCV = {Int. J. Comput. Vis.})

@String(CVPR= {IEEE Conf. Comput. Vis. Pattern Recog.})

@String(ICCV= {Int. Conf. Comput. Vis.})

@String(ECCV= {Eur. Conf. Comput. Vis.})

@String(TMM  = {IEEE Trans. Multimedia})

@String(ACMMM= {ACM Int. Conf. Multimedia})

@String(AAAI = {AAAI})

@String(TPAMI  = {IEEE TPAMI})

@String(IJCV  = {IJCV})

@String(CVPR  = {CVPR})

@String(ICCV  = {ICCV})

@String(ECCV  = {ECCV})

@String(NeurIPS  = {NeurIPS})

@String(TMM   =	{IEEE TMM})

@String(ACMMM = {ACM MM})

@String(WACV = {WACV})

@article{MeViSv2,
  title={{MeViS}: A Multi-Modal Dataset for Referring Motion Expression Video Segmentation},
  author={Ding, Henghui and Liu, Chang and He, Shuting and Ying, Kaining and Jiang, Xudong and Loy, Chen Change and Jiang, Yu-Gang},
  journal=TPAMI,
  year={2025},
  publisher={IEEE}
}

@inproceedings{MeViS,
  title={{MeViS}: A Large-scale Benchmark for Video Segmentation with Motion Expressions},
  author={Ding, Henghui and Liu, Chang and He, Shuting and Jiang, Xudong and Loy, Chen Change},
  booktitle=ICCV,
  year={2023}
}

@inproceedings{xu2017deep,
  title={Deep image matting},
  author={Xu, Ning and Price, Brian and Cohen, Scott and Huang, Thomas},
  booktitle=CVPR,
  xxxxpages={2970--2979},
  year={2017}
}

@inproceedings{liu2021tripartite,
  title={Tripartite information mining and integration for image matting},
  author={Liu, Yuhao and Xie, Jiake and Shi, Xiao and Qiao, Yu and Huang, Yujie and Tang, Yong and Yang, Xin},
  booktitle=ICCV,
  xxxxpages={7555--7564},
  year={2021}
}

@inproceedings{park2022matteformer,
  title={Matteformer: Transformer-based image matting via prior-tokens},
  author={Park, GyuTae and Son, SungJoon and Yoo, JaeYoung and Kim, SeHo and Kwak, Nojun},
  booktitle=CVPR,
  xxxxpages={11696--11706},
  year={2022}
}

@inproceedings{li2024drip,
  title={Drip: Unleashing diffusion priors for joint foreground and alpha prediction in image matting},
  author={Li, Xiaodi and Yang, Zongxin and Quan, Ruijie and Yang, Yi},
  booktitle=NeurIPS,
  xxxxpages={79868--79888},
  year={2024}
}

@inproceedings{hu2024diffusion,
  title={Diffusion for natural image matting},
  author={Hu, Yihan and Lin, Yiheng and Wang, Wei and Zhao, Yao and Wei, Yunchao and Shi, Humphrey},
  booktitle=ECCV,
  xxxxpages={181--199},
  year={2024},
  organization={Springer}
}

@inproceedings{ke2022modnet,
  title={Modnet: Real-time trimap-free portrait matting via objective decomposition},
  author={Ke, Zhanghan and Sun, Jiayu and Li, Kaican and Yan, Qiong and Lau, Rynson WH},
  booktitle=AAAI,
  volume={36},
  xxxxpages={1140--1147},
  year={2022}
}

@inproceedings{park2023mask,
  title={Mask-guided matting in the wild},
  author={Park, Kwanyong and Woo, Sanghyun and Oh, Seoung Wug and Kweon, In So and Lee, Joon-Young},
  booktitle=CVPR,
  xxxxpages={1992--2001},
  year={2023}
}

@article{yao2024matte,
  title={Matte anything: Interactive natural image matting with segment anything model},
  author={Yao, Jingfeng and Wang, Xinggang and Ye, Lang and Liu, Wenyu},
  journal={Image and Vision Computing},
  volume={147},
  xxxxpages={105067},
  year={2024},
  publisher={Elsevier}
}

@inproceedings{li2024matting,
  title={Matting anything},
  author={Li, Jiachen and Jain, Jitesh and Shi, Humphrey},
  booktitle=CVPR,
  xxxxpages={1775--1785},
  year={2024}
}

@inproceedings{huang2023end,
  title={End-to-end video matting with trimap propagation},
  author={Huang, Wei-Lun and Lee, Ming-Sui},
  booktitle=CVPR,
  xxxxpages={14337--14347},
  year={2023}
}

@inproceedings{huynh2024maggie,
  title={Maggie: Masked guided gradual human instance matting},
  author={Huynh, Chuong and Oh, Seoung Wug and Shrivastava, Abhinav and Lee, Joon-Young},
  booktitle=CVPR,
  xxxxpages={3870--3879},
  year={2024}
}

@inproceedings{yang2025matanyone,
  title={MatAnyone: Stable video matting with consistent memory propagation},
  author={Yang, Peiqing and Zhou, Shangchen and Zhao, Jixin and Tao, Qingyi and Loy, Chen Change},
  booktitle=CVPR,
  xxxxpages={7299--7308},
  year={2025}
}

@inproceedings{lin2022robust,
  title={Robust high-resolution video matting with temporal guidance},
  author={Lin, Shanchuan and Yang, Linjie and Saleemi, Imran and Sengupta, Soumyadip},
  booktitle=WACV,
  xxxxpages={238--247},
  year={2022}
}

@inproceedings{zhang2021attention,
  title={Attention-guided temporally coherent video object matting},
  author={Zhang, Yunke and Wang, Chi and Cui, Miaomiao and Ren, Peiran and Xie, Xuansong and Hua, Xian-Sheng and Bao, Hujun and Huang, Qixing and Xu, Weiwei},
  booktitle=ACMMM,
  xxxxpages={5128--5137},
  year={2021}
}

@inproceedings{li2021privacy,
  title={Privacy-preserving portrait matting},
  author={Li, Jizhizi and Ma, Sihan and Zhang, Jing and Tao, Dacheng},
  booktitle=ACMMM,
  xxxxpages={3501--3509},
  year={2021}
}

@inproceedings{hou2019context,
  title={Context-aware image matting for simultaneous foreground and alpha estimation},
  author={Hou, Qiqi and Liu, Feng},
  booktitle=ICCV,
  xxxxpages={4130--4139},
  year={2019}
}

@inproceedings{lin2021real,
  title={Real-time high-resolution background matting},
  author={Lin, Shanchuan and Ryabtsev, Andrey and Sengupta, Soumyadip and Curless, Brian L and Seitz, Steven M and Kemelmacher-Shlizerman, Ira},
  booktitle=CVPR,
  xxxxpages={8762--8771},
  year={2021}
}

@inproceedings{yu2021cascade,
  title={Cascade image matting with deformable graph refinement},
  author={Yu, Zijian and Li, Xuhui and Huang, Huijuan and Zheng, Wen and Chen, Li},
  booktitle=ICCV,
  xxxxpages={7167--7176},
  year={2021}
}

@inproceedings{yu2021high,
  title={High-resolution deep image matting},
  author={Yu, Haichao and Xu, Ning and Huang, Zilong and Zhou, Yuqian and Shi, Humphrey},
  booktitle=AAAI,
  xxxxpages={3217--3224},
  year={2021}
}

@inproceedings{li2023referring,
  title={Referring image matting},
  author={Li, Jizhizi and Zhang, Jing and Tao, Dacheng},
  booktitle=CVPR,
  xxxxpages={22448--22457},
  year={2023}
}

@article{yang2020smart,
  title={Smart scribbles for image matting},
  author={Yang, Xin and Qiao, Yu and Chen, Shaozhe and He, Shengfeng and Yin, Baocai and Zhang, Qiang and Wei, Xiaopeng and Lau, Rynson WH},
  journal={ACM Transactions on Multimedia Computing, Communications, and Applications (TOMM)},
  volume={16},
  number={4},
  xxxxpages={1--21},
  year={2020},
  publisher={ACM New York, NY, USA}
}

@article{ravi2024sam,
  title={Sam 2: Segment anything in images and videos},
  author={Ravi, Nikhila and Gabeur, Valentin and Hu, Yuan-Ting and Hu, Ronghang and Ryali, Chaitanya and Ma, Tengyu and Khedr, Haitham and R{\"a}dle, Roman and Rolland, Chloe and Gustafson, Laura and others},
  journal={arXiv preprint arXiv:2408.00714},
  year={2024}
}

@inproceedings{sun2021deep,
  title={Deep video matting via spatio-temporal alignment and aggregation},
  author={Sun, Yanan and Wang, Guanzhi and Gu, Qiao and Tang, Chi-Keung and Tai, Yu-Wing},
  booktitle=CVPR,
  xxxxpages={6975--6984},
  year={2021}
}

@inproceedings{li2024vmformer,
  title={Vmformer: End-to-end video matting with transformer},
  author={Li, Jiachen and Goel, Vidit and Ohanyan, Marianna and Navasardyan, Shant and Wei, Yunchao and Shi, Humphrey},
  booktitle=WACV,
  xxxxpages={6678--6687},
  year={2024}
}

@inproceedings{li2023videomatt,
  title={VideoMatt: A simple baseline for accessible real-time video matting},
  author={Li, Jiachen and Ohanyan, Marianna and Goel, Vidit and Navasardyan, Shant and Wei, Yunchao and Shi, Humphrey},
  booktitle=CVPR,
  xxxxpages={2177--2186},
  year={2023}
}

@inproceedings{seong2022one,
  title={One-trimap video matting},
  author={Seong, Hongje and Oh, Seoung Wug and Price, Brian and Kim, Euntai and Lee, Joon-Young},
  booktitle=ECCV,
  xxxxpages={430--448},
  year={2022},
  organization={Springer}
}

@article{li2022bridging,
  title={Bridging composite and real: towards end-to-end deep image matting},
  author={Li, Jizhizi and Zhang, Jing and Maybank, Stephen J and Tao, Dacheng},
  journal=IJCV,
  volume={130},
  number={2},
  xxxxpages={246--266},
  year={2022},
  publisher={Springer}
}

@article{li2021deep,
  title={Deep automatic natural image matting},
  author={Li, Jizhizi and Zhang, Jing and Tao, Dacheng},
  journal={arXiv preprint arXiv:2107.07235},
  year={2021}
}

@inproceedings{qiao2020attention,
  title={Attention-guided hierarchical structure aggregation for image matting},
  author={Qiao, Yu and Liu, Yuhao and Yang, Xin and Zhou, Dongsheng and Xu, Mingliang and Zhang, Qiang and Wei, Xiaopeng},
  booktitle=CVPR,
  xxxxpages={13676--13685},
  year={2020}
}

@article{yang2022exploring,
  title={Exploring the Interactive Guidance for Unified and Effective Image Matting},
  author={Yang, Dinghao and Wang, Bin and Li, Weijia and Lin, YiQi and He, Conghui},
  journal={arXiv preprint arXiv:2205.08324},
  year={2022}
}

@InProceedings{Liu_2021_WACV,
    author    = {Liu, Chang and Ding, Henghui and Jiang, Xudong},
    title     = {Towards Enhancing Fine-Grained Details for Image Matting},
    booktitle = WACV,
    year      = {2021},
    xxxxpages     = {385-393}
}

@article{loshchilov2017decoupled,
  title={Decoupled weight decay regularization},
  author={Loshchilov, Ilya and Hutter, Frank},
  journal={arXiv preprint arXiv:1711.05101},
  year={2017}
}

@inproceedings{lin2023adaptive,
  title={Adaptive human matting for dynamic videos},
  author={Lin, Chung-Ching and Wang, Jiang and Luo, Kun and Lin, Kevin and Li, Linjie and Wang, Lijuan and Liu, Zicheng},
  booktitle=CVPR,
  xxxxpages={10229--10238},
  year={2023}
}

@inproceedings{wang2023video,
  title={Video object matting via hierarchical space-time semantic guidance},
  author={Wang, Yumeng and Xu, Bo and Li, Ziwen and Huang, Han and Lu, Cheng and Guo, Yandong},
  booktitle=WACV,
  xxxxpages={5120--5129},
  year={2023}
}

@article{zhang2025object,
  title={Object-Aware Video Matting with Cross-Frame Guidance},
  author={Zhang, Huayu and Wu, Dongyue and Shao, Yuanjie and Sang, Nong and Gao, Changxin},
  journal={arXiv preprint arXiv:2503.01262},
  year={2025}
}

@inproceedings{sharma2020alphanet,
  title={AlphaNet: An attention guided deep network for automatic image matting},
  author={Sharma, Rishab and Deora, Rahul and Vishvakarma, Anirudha},
  booktitle={International Conference on Omni-layer Intelligent Systems (COINS)},
  xxxxpages={1--8},
  year={2020},
  organization={IEEE}
}

@article{zhou2021semantic,
  title={Semantic-guided Automatic Natural Image Matting with Trimap Generation Network and Light-weight Non-local Attention},
  author={Zhou, Yuhongze and Zhou, Liguang and Lam, Tin Lun and Xu, Yangsheng},
  journal={arXiv preprint arXiv:2103.17020},
  year={2021}
}

@inproceedings{cheng2022masked,
  title={Masked-attention mask transformer for universal image segmentation},
  author={Cheng, Bowen and Misra, Ishan and Schwing, Alexander G and Kirillov, Alexander and Girdhar, Rohit},
  booktitle=CVPR,
  xxxxpages={1290--1299},
  year={2022}
}

@inproceedings{yu2021mask,
  title={Mask guided matting via progressive refinement network},
  author={Yu, Qihang and Zhang, Jianming and Zhang, He and Wang, Yilin and Lin, Zhe and Xu, Ning and Bai, Yutong and Yuille, Alan},
  booktitle=CVPR,
  xxxxpages={1154--1163},
  year={2021}
}

@article{deora2021salient,
  title={Salient image matting},
  author={Deora, Rahul and Sharma, Rishab and Raj, Dinesh Samuel Sathia},
  journal={arXiv preprint arXiv:2103.12337},
  year={2021}
}

@article{dai2022enabling,
  title={Enabling trimap-free image matting with a frequency-guided saliency-aware network via joint learning},
  author={Dai, Linhui and Song, Xiang and Liu, Xiaohong and Li, Chengqi and Shi, Zhihao and Chen, Jun and Brooks, Martin},
  journal=TMM,
  volume={25},
  xxxxpages={4868--4879},
  year={2022},
  publisher={IEEE}
}

@inproceedings{zhong2024lightweight,
  title={Lightweight portrait matting via regional attention and refinement},
  author={Zhong, Yatao and Zharkov, Ilya},
  booktitle=WACV,
  xxxxpages={4158--4167},
  year={2024}
}

@article{ding2022deep,
  title={Deep interactive image matting with feature propagation},
  author={Ding, Henghui and Zhang, Hui and Liu, Chang and Jiang, Xudong},
  journal={IEEE Transactions on Image Processing},
  volume={31},
  xxxxpages={2421--2432},
  year={2022},
  publisher={IEEE}
}

@inproceedings{wei2021improved,
  title={Improved image matting via real-time user clicks and uncertainty estimation},
  author={Wei, Tianyi and Chen, Dongdong and Zhou, Wenbo and Liao, Jing and Zhao, Hanqing and Zhang, Weiming and Yu, Nenghai},
  booktitle=CVPR,
  xxxxpages={15374--15383},
  year={2021}
}

@article{tan2016novel,
  title={A novel image matting method using sparse manual clicks},
  author={Tan, Guanghua and Chen, Hui and Qi, Jun},
  journal={Multimedia Tools and Applications},
  volume={75},
  number={17},
  xxxxpages={10213--10225},
  year={2016},
  publisher={Springer}
}

@article{liu2025enhancing,
  title={Enhancing Image Matting in Real-World Scenes with Mask-Guided Iterative Refinement},
  author={Liu, Rui},
  journal={arXiv preprint arXiv:2502.17093},
  year={2025}
}

@article{liu2023end,
  title={End-to-end human instance matting},
  author={Liu, Qinglin and Zhang, Shengping and Meng, Quanling and Zhong, Bineng and Liu, Peiqiang and Yao, Hongxun},
  journal={IEEE Transactions on Circuits and Systems for Video Technology},
  volume={34},
  number={4},
  xxxxpages={2633--2647},
  year={2023},
  publisher={IEEE}
}

@inproceedings{zhang2019late,
  title={A late fusion cnn for digital matting},
  author={Zhang, Yunke and Gong, Lixue and Fan, Lubin and Ren, Peiran and Huang, Qixing and Bao, Hujun and Xu, Weiwei},
  booktitle=CVPR,
  xxxxpages={7469--7478},
  year={2019}
}

@inproceedings{sun2022human,
  title={Human instance matting via mutual guidance and multi-instance refinement},
  author={Sun, Yanan and Tang, Chi-Keung and Tai, Yu-Wing},
  booktitle=CVPR,
  xxxxpages={2647--2656},
  year={2022}
}

@inproceedings{sun2023ultrahigh,
  title={Ultrahigh resolution image/video matting with spatio-temporal sparsity},
  author={Sun, Yanan and Tang, Chi-Keung and Tai, Yu-Wing},
  booktitle=CVPR,
  xxxxpages={14112--14121},
  year={2023}
}

@article{ding2025mosev2,
  title={{MOSEv2}: A more challenging dataset for video object segmentation in complex scenes},
  author={Ding, Henghui and Ying, Kaining and Liu, Chang and He, Shuting and Jiang, Xudong and Jiang, Yu-Gang and Torr, Philip HS and Bai, Song},
  journal={arXiv preprint arXiv:2508.05630},
  year={2025}
}

@inproceedings{ding2023mose,
  title={{MOSE}: A new dataset for video object segmentation in complex scenes},
  author={Ding, Henghui and Liu, Chang and He, Shuting and Jiang, Xudong and Torr, Philip HS and Bai, Song},
  booktitle=ICCV,
  xxxxpages={20224--20234},
  year={2023}
}

@article{lim2026videomama,
  title={VideoMaMa: Mask-Guided Video Matting via Generative Prior},
  author={Lim, Sangbeom and Oh, Seoung Wug and Huang, Jiahui and Yoon, Heeji and Kim, Seungryong and Lee, Joon-Young},
  journal={arXiv preprint arXiv:2601.14255},
  year={2026}
}

@article{yang2025matanyone2,
  title={MatAnyone 2: Scaling Video Matting via a Learned Quality Evaluator},
  author={Yang, Peiqing and Zhou, Shangchen and Hao, Kai and Tao, Qingyi},
  journal={arXiv preprint arXiv:2512.11782},
  year={2025}
}

@article{carion2025sam,
  title={Sam 3: Segment anything with concepts},
  author={Carion, Nicolas and Gustafson, Laura and Hu, Yuan-Ting and Debnath, Shoubhik and Hu, Ronghang and Suris, Didac and Ryali, Chaitanya and Alwala, Kalyan Vasudev and Khedr, Haitham and Huang, Andrew and others},
  journal={arXiv preprint arXiv:2511.16719},
  year={2025}
}

@misc{celebahairmaskhq,
  title        = {CelebAHairMask-HQ},
  author       = {Gao, Zhihan},
  howpublished = {\url{https://github.com/cpuimage/CelebAHairMask-HQ}},
  note         = {Accessed: 2026-05-29},
  year = {2025},
}

\end{document}